\documentclass[10pt,twocolumn,letterpaper]{article}

\usepackage{cvpr}
\usepackage{times}
\usepackage{epsfig}
\usepackage{graphicx}
\usepackage{amsmath}
\usepackage{amssymb}

\iftrue     
\newcommand{\dimpp}[1]{\textcolor{blue}{[DP: #1]}}
\newcommand{\jasper}[1]{\textcolor{magenta}{[JU: #1]}}
\newcommand{\vitto}[1]{\textcolor{red}{[VF: #1]}}
\newcommand{\frank}[1]{\textcolor{cyan}{[FK: #1]}}
\else
\newcommand{\dimpp}[1]{\textcolor{blue}{\noindent}}
\newcommand{\jasper}[1]{\textcolor{magenta}{\noindent}}
\newcommand{\vitto}[1]{\textcolor{red}{\noindent}}
\newcommand{\frank}[1]{\textcolor{cyan}{\noindent}}
\fi

\newcommand{\mypar}[1]{\vspace{-5mm}\paragraph{#1}}

\DeclareMathAlphabet{\mathpzc}{T1}{pzc}{m}{n}
\newlength{\halfwidth}
\newlength{\fullwidth}
\newlength{\tikzimgheight}
\newlength{\tikzimgwidth}

\usepackage{ifthen}

\usepackage[pagebackref=true,breaklinks=true,letterpaper=true,colorlinks,bookmarks=false]{hyperref}

\cvprfinalcopy % *** Uncomment this line for the final submission

\def\cvprPaperID{2827} % *** Enter the CVPR Paper ID here

\ifcvprfinal\fi
\begin{document}

%%%%%%%%% TITLE
\title{Training object class detectors with click supervision}

\author{\hspace{-0.8cm} Dim P. Papadopoulos\textsuperscript{1} \hspace{1.2cm} Jasper R. R. Uijlings\textsuperscript{2} \hspace{1.0cm} Frank Keller\textsuperscript{1} \hspace{1.7cm} Vittorio Ferrari\textsuperscript{1,2} \\
{\tt\small \hspace{-0.8cm} dim.papadopoulos@ed.ac.uk \hspace{0.4cm} jrru@google.com \hspace{0.15cm} keller@inf.ed.ac.uk \hspace{0.2cm} vferrari@inf.ed.ac.uk}\\
\textsuperscript{1}University of Edinburgh  \hspace{1cm} \textsuperscript{2}Google Research
}

\maketitle	

\begin{abstract}
\vspace{-0.08cm}

Training object class detectors typically requires a large set of images with objects annotated by bounding boxes. However, manually drawing bounding boxes is very time consuming.
In this paper we greatly reduce annotation time by proposing center-click annotations: we ask annotators to click on the center of an imaginary bounding box which tightly encloses the object instance.
We then incorporate these clicks into existing Multiple Instance Learning techniques for weakly supervised object localization, to jointly localize object bounding boxes over all training images.
Extensive experiments on PASCAL VOC 2007 and MS COCO show that:
(1) our scheme delivers high-quality detectors, performing substantially better than those produced by weakly supervised techniques, with a modest extra annotation effort; 
(2) these detectors in fact perform in a range close to those trained from manually drawn bounding boxes;
(3) as the center-click task is very fast, our scheme reduces total annotation time by $9\times$ to $18\times$.
% \dimpp{It's 9.4-18.7. (35 seconds vs. 1.87 seconds)}

%
%\vitto{we should reword the tension between VOC and COCO, and between real clicks and simulations; I'll revisit after a full read}
%We crowd-source center-clicks for PASCAL VOC 2007 and demonstrate that
%(a)~our proposed training scheme leads to high quality bounding box annotations, while reducing the total human annotation time by $11\times$ to $22\times$.
%%% it's "by a factor of n" or "by n times", not both!
%On PASCAL VOC 2007 and MS~COCO we show that 
%(b) it delivers object detectors performing substantially better than existing state-of-the-art weakly supervised object localization techniques, with a modest extra annotation effort;
%\vitto{the 3 points where very flat before, all just saying 'good'; now I'm trying to make one bullet about the BBs in the trn set, one about the output detectors' perf on the test set}
%(c) given the same annotation time, our scheme outperforms all existing forms of supervision for object detection.
\end{abstract}

\vspace{-0.2cm}
\section{Introduction}
\label{sec:intro}

%\dimpp{ should agree on: (1) center or middle? (2) click or point? ...}
%\vitto{I vote for: (1) center; (2) no preference}

How can we train high-quality computer vision models with minimal human annotation effort?
% VF: I remove the following point, also from the abstract. It does not work in our paper, as we annotate datasets of the same size as in the pre-CNN era; in fact, modern CNN object detectors take the SAME AMOUNT OF BBs as pre-CNN detectors! They are data hungry 'only' in the image-level label pre-training stage; it is important not to send the reviewer up the wrong expectation avenue
%This question is important since the best models use data-hungry neural networks.
Obtaining training data is especially costly for object class detection, the task of detecting all instances of a given
object class in an image. Typically, detectors are trained under full supervision, which
requires manually drawing tight object bounding boxes in a large number of training images.
This takes time: annotating the popular ILSVRC dataset~\cite{russakovsky15ijcv} required about 35s
per bounding box, using a crowd-sourcing technique optimized for efficient bounding box
annotation~\cite{su12aaai} (more details in Sec. \ref{sec:relwork}).
% We further discuss the time to annotate a bounding box in Sec.~\ref{sec:relwork}. }

Object detectors can also be trained
under weak supervision using only image-level labels. While this is substantially cheaper, the resulting
detectors typically deliver only about half the accuracy of their fully supervised
counterparts~\cite{bilen14bmvc,bilen15cvpr,bilen16cvpr,cinbis15pami,deselaers10eccv,kantorov16eccv,russakovsky12eccv,siva11iccv,song14icml,song14nips,wang15tip}.
In this paper, we aim to minimize human annotation effort while producing high-quality detectors. To this end we
propose annotating objects by clicking on their center.

%\jasper{Frank, can you please contribute to the next paragraph?}
%\frank{I have added citations, and removed ``(3)~is a low-cognitive human
%task'', which probably means: clicking is doesn't require a lot of
%cognitive effort. This is probably correct, but I don't think there are any studies
%showing this, so we can't really claim it.}
Clicking on an object can be seen as the human-computer-interaction equivalent of pointing to an
object. Pointing is a natural way for humans to
communicate that emerges early during cognitive
development~\cite{tomasello07childdevelopment}. Human pointing
behavior is well-understood in human-computer interaction, and can be
modeled mathematically~\cite{Soukoreff:ea:04}. For the purpose of
image annotation, clicking on an object is therefore a natural choice.
Clicking offers several advantages over other ways to annotate bounding boxes:
(1)~is substantially faster than drawing bounding boxes~\cite{su12aaai},
(2)~requires little instructions or annotator training compared to drawing~\cite{su12aaai} or verifying bounding
boxes~\cite{papadopoulos16cvpr,russakovsky15cvpr,su12aaai}, because it is a task that comes natural to humans, 
(3)~can be performed using a simple annotation interface (unlike bounding box drawing~\cite{su12aaai}), and requires no specialized hardware (unlike eye-tracking~\cite{papadopoulos14eccv}). Note that the scheme we propose does not require a human-in-the-loop setup~\cite{deng13cvpr,papadopoulos16cvpr,parkash12eccv,vijayanarasimhan14ijcv,jain16cvpr}: clicks can be acquired separately, independently of the detector training framework used.
%\frank{Removed the self-citation here; this point applies to all HitL
%  setups, not just to ours.}

% Clicking on an object can be seen as the human-computer-interaction-equivalent of pointing to an
% object. Pointing again is a natural way for humans to communicate~\cite{tomasello07, ???}.
% Therefore, clicking on objects is a natural way to annotate images. Intuitively, clicking (1) is
% fast, (2) requires few instructions for the annotator, (3) is a low-cognitive human task, and (4)
% requires only a simple annotation interface. 

\begin{figure*}[t]
\vspace{-.6cm}
\center
\includegraphics[width=\linewidth]{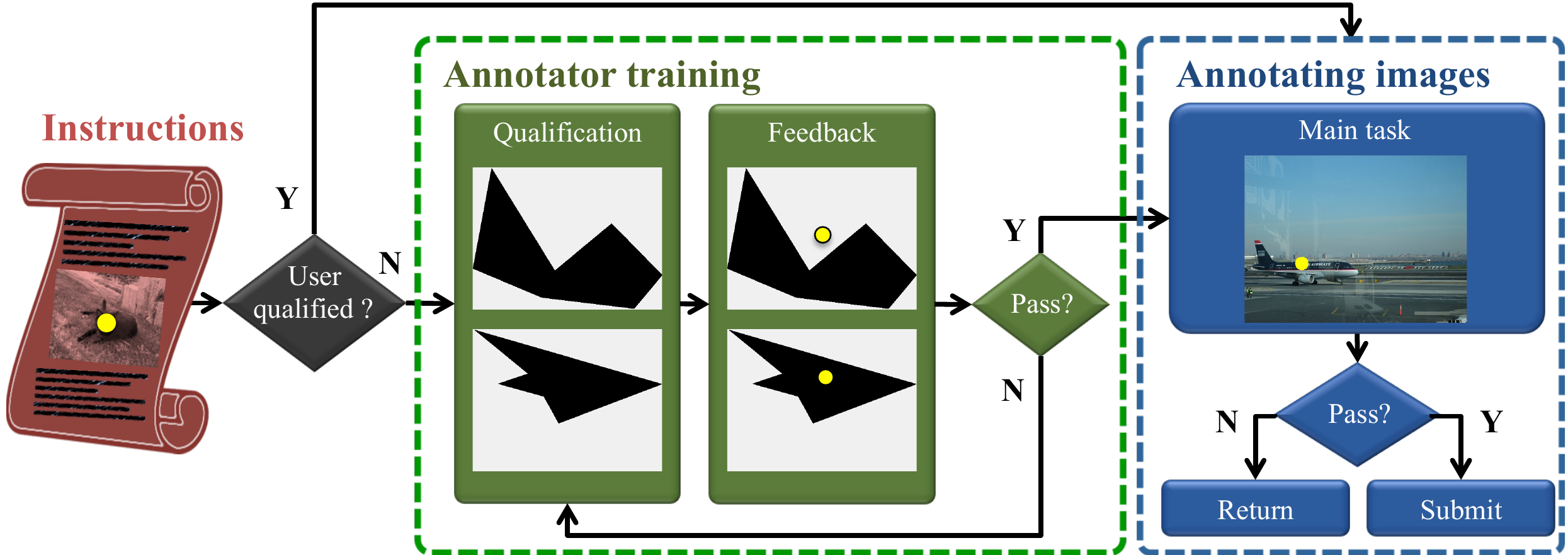}
\vspace{-.2cm}
\caption{\small \textbf{The workflow of our crowd-sourcing framework for collecting click annotations.} The annotators read a set of instructions and then go through an interactive training stage that consists of a simple qualification test based on synthetic polygons. After completing it, they receive a detailed feedback on how well they performed. Annotators who successfully pass the qualification test can proceed to the annotation stage. In case of failure, they can repeat the test as many times as they want.}
\vspace{-.4cm}
\label{fig:mturkflow}
\end{figure*}

% all about clicks and object properties we can derive from them
%In this paper 
Given an image known to contain a certain object class,
we ask annotators to click on the center of an imaginary bounding box enclosing the object ({\em center-click} annotations). These clicks provide reliable anchor points for the full bounding box, as they provide an estimate of its center.
Moreover, we can also ask two different annotators to provide center-clicks on the same object. As their errors are independent, we can obtain a more accurate estimate of the object center by averaging their click positions.
Interestingly, given the two clicks, we can even estimate the {\em size} of the object, by exploiting a correlation between the object size and the distance of the click to the true center (error). As the errors are independent, the distance between the two clicks increases with object size. This enables to estimate size based on the distance between the clicks.
As a novel component of our crowd-sourcing protocol, we introduce a stage to train the annotators based on synthetic polygons. This enables generating an arbitrarily large set of training questions without using any manually drawn bounding box.
Moreover, we derive models of the annotator error directly from this polygon stage, and use them later to estimate object size in real images.

% all about incorporating clicks into MIL
We incorporate these clicks into a reference Multiple Instance Learning (MIL) framework which was originally designed for weakly supervised object detection~\cite{cinbis15pami}. It jointly localizes object bounding boxes over all training images of an object class. It iteratively alternates between re-training the detector and re-localizing objects.
We use the center-clicks in the re-localization phase, % and also in initialization, but hey, too much detail for the intro
to promote selecting % producing; I'm surfing on the wording for object proposals
bounding boxes % VF: I'd like to say selecting object proposals, but maybe too much detail for the intro
compatible with the object center and size estimated based on the clicks.

% result shoutout
Based on extensive experiments with crowd-sourced center-clicks on Amazon Mechanical Turk for PASCAL VOC 2007 and simulations on MS COCO, we demonstrate that:
(1) our scheme incorporating center-click into MIL delivers better bounding boxes on the training set. In turn, this lead to high-quality detectors, performing substantially better than those produced by weakly supervised techniques, with a modest extra annotation effort (less than 4h on the entire PASCAL VOC 2007 trainval);
(2) these detectors in fact perform in a range close to those trained from manually drawn bounding boxes;
(3) as the center-click task is very fast, our scheme reduces total annotation time by $9\times$ (one click) to $18\times$ (two clicks);
(4) given the same human annotation budget, our scheme outperforms the recent human verification scheme~\cite{papadopoulos16cvpr}, which was already very efficient.

\section{Related work}
\label{sec:relwork}

\paragraph{Time to draw a bounding box.}

The time required to draw a bounding box varies depending on several factors, including the desired quality of the boxes and the particular crowdsourcing protocol used.
In this paper, as an authoritative reference we use the protocol of~\cite{su12aaai} which was used to annotate ILSVRC~\cite{russakovsky15ijcv}.
It was designed to produce high-quality bounding boxes with minimal
human annotation time on Amazon Mechanical Turk, a popular crowdsourcing platform.
They report the following median times for annotating an object class in an image~\cite{su12aaai}:
25.5s for drawing one box,
9.0s for verifying its quality,
and 7.8s for checking whether there are other objects of the same class yet to be annotated (in which case the process repeats).
Since we only consider localizing one object per class per image, we use $25.5s + 9.0s = 34.5s$ as the reference time for manually annotating a high-quality bounding box.
This is a conservative estimate: when taking into account
that some boxes are rejected in the second step and need to be re-drawn multiple times until they are correct, the median time increases to 55s. If we use average times instead of medians, the cost raises further to 117s.
%\dimpp{why did we cut the 1 decimal point from these 2 numbers? 55.48 and 116.9}

We use 34.5s as reference both for PASCAL VOC~\cite{everingham10ijcv}, which has objects of
comparable difficulty to ILSVRC~\cite{russakovsky15ijcv}, and for COCO~\cite{lin14eccv}, which is more difficult. Both datasets have high-quality bounding boxes, which we use as reference for comparisons to our method.

\paragraph{Weakly-supervised object localization (WSOL).}

These methods are trained from a set of images labeled only as containing a certain object class, without being given the location of the objects~\cite{bilen14bmvc,bilen15cvpr,bilen16cvpr,cinbis14cvpr,deselaers10eccv,kantorov16eccv,russakovsky12eccv,siva11iccv,song14icml,song14nips,wang15tip}. 
The goal is to localize the objects in these training images while learning an object detector for localizing instances in new test images.
Recent work on WSOL~\cite{bilen14bmvc,bilen15cvpr,bilen16cvpr,cinbis14cvpr,kantorov16eccv,song14icml,song14nips,wang15tip} has shown remarkable progress thanks to Convolutional Neural Nets (CNNs~\cite{girshick14cvpr,krizhevsky12nips}). However, learning a detector without location annotations is difficult and performance is generally about half that of their fully supervised counterparts
%(typically about half the mAP of the same detection model trained on the same images but with manual bounding-box annotation) 
~\cite{bilen14bmvc,bilen15cvpr,bilen16cvpr,cinbis14cvpr,deselaers10eccv,kantorov16eccv,russakovsky12eccv,siva11iccv,song14icml,song14nips,wang15tip}.

WSOL is often addressed as a Multiple Instance Learning (MIL)
problem~\cite{bilen14bmvc,cinbis14cvpr,deselaers10eccv,dietterich97ai,shi12bmvc,siva11iccv,song14icml,song14nips}. In this paper, we use MIL as our basis and augment it with center-click supervision.

% WSOL is often addressed as a Multiple Instance Learning (MIL) problem~\cite{bilen14bmvc,cinbis14cvpr,deselaers10eccv,dietterich97ai,shi12bmvc,siva11iccv,song14icml,song14nips}.
% Images are treated as bags of windows (object proposals~\cite{alexe10cvpr,dollar14eccv,uijlings13ijcv}). A negative image contains only negative instances, while a positive one contains at least one positive instance, mixed in with a majority of negatives.  The goal is to find the true positives instances from which to learn a window classifier for the object class. This typically happens by iteratively alternating between re-training the object detector given the current selection of positive instances and re-localising instances in the positive images using the current object detector. In this paper, we build in a standard MIL WSOL model and we improve the re-localization step using click supervision. Our proposed framework leads to substantially better object localizations and is able to train high quality object detectors with a modest extra annotation cost.

\mypar{Click supervision.}

Click annotation schemes have been used in part-based detection to annotate part locations of an object~\cite{branson11iccv,wah11iccv}, and in human pose estimation to annotate key-points of human body parts~\cite{johnson10bmvc,Ramanan06-fixed,sapp13cvpr}.
Click supervision has also been used to reduce the annotation time for semantic segmentation~\cite{bearman16eccv,jain16hcomp,bell15cvpr,wang14cviu}. Recently, Bearman et al.~\cite{bearman16eccv} collected clicks by asking the annotators to click anywhere on a target object. In Sec.~\ref{sec:resultsVOC}, we show that our center-click annotations outperforms these click-anywhere annotations for object class detection.
Finally, Mettes et al.~\cite{mettes16eccv} proposed to annotate actions in videos with click annotations. 
Our work also offers other new elements over the above works, e.g. estimating object area from two clicks and training annotators with synthetic polygons.
% VF: also do it for obj det, not sem segm, so 'selecting proposals' fmw

\mypar{Other ways to reduce annotation cost.}
Researchers tried to learn object class detectors from videos, where the spatio-temporal continuity facilitates object localization~\cite{kalogeiton16pami,leistner11cvpr,prest12cvpr,singh16cvpr,Tang2013}. 
An alternative direction is transfer learning, where an appearance model for a new class is learned from bounding box annotations on examples of related classes~\cite{Aytar11iccv,fei2007CVIU,guillaumin12cvpr,hoffman14nips,kuettel12eccv,lampert:cvpr09,rochan15cvpr}.
Eye-tracking data can be seen as another type of pointing to an object. Such data have been used as a weak supervisory signal to localize objects on images~\cite{mathe13nips,papadopoulos14eccv} or videos~\cite{karthikeyan2015cvpr,mathe12eccv}.

Recently, Papadopoulos et al.~\cite{papadopoulos16cvpr} proposed a very efficient 
%human-in-the-loop 
framework for training object class detectors that only requires humans to verify bounding boxes produced by the learning algorithm. We compare with~\cite{papadopoulos16cvpr} in Sec.~\ref{sec:results}.

\section{Crowd-sourcing clicks}
\label{sec:mturk}

\begin{figure}[t]
\vspace{-.6cm}
\center
\begin{tabular}{c@{}c@{}c@{}c@{}c@{}}
{aeroplane} &  & {bicycle} &  & {bus} \\
\includegraphics[width=0.32\linewidth]{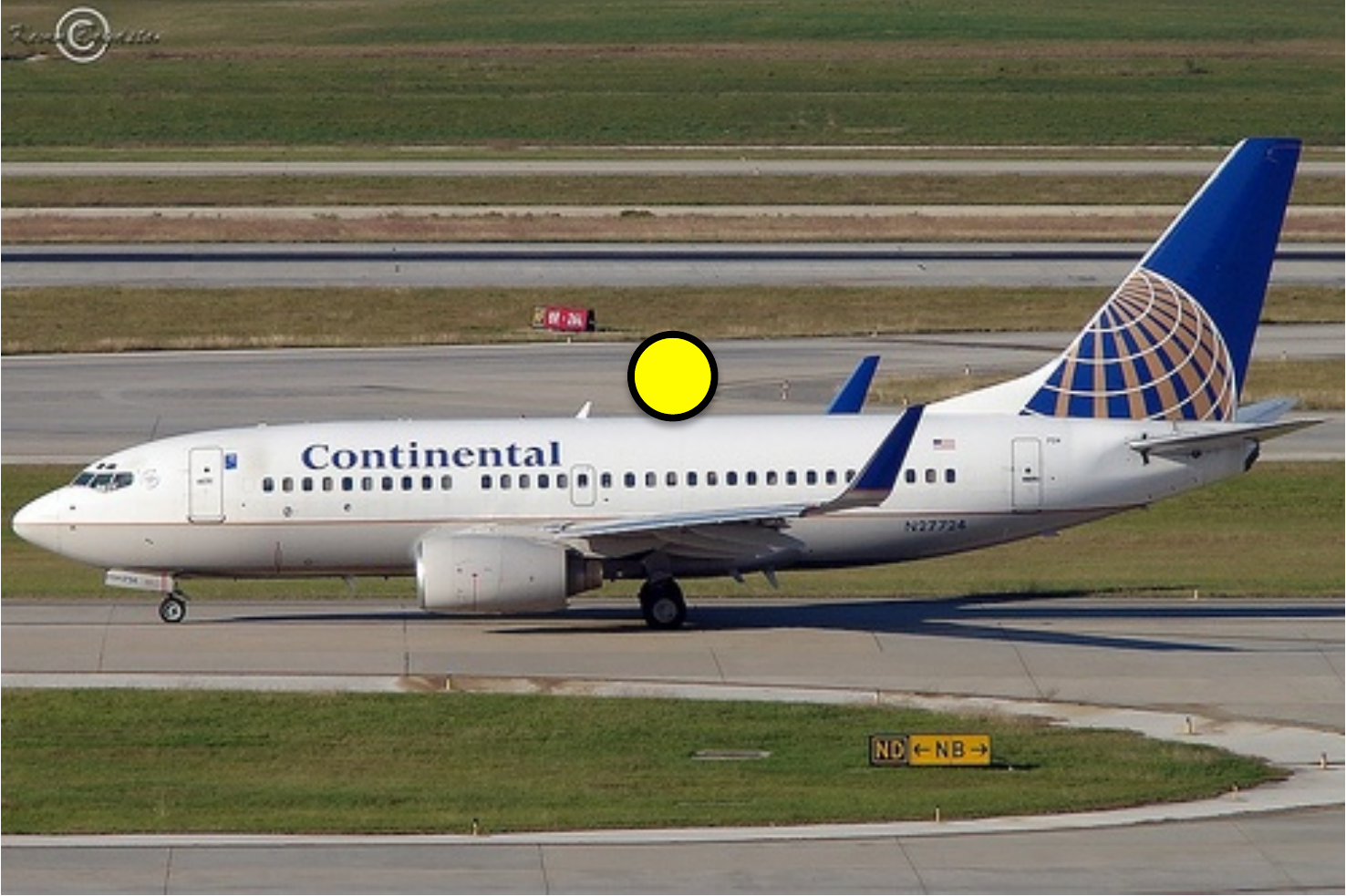}&
\textit{ } &
\includegraphics[width=0.32\linewidth]{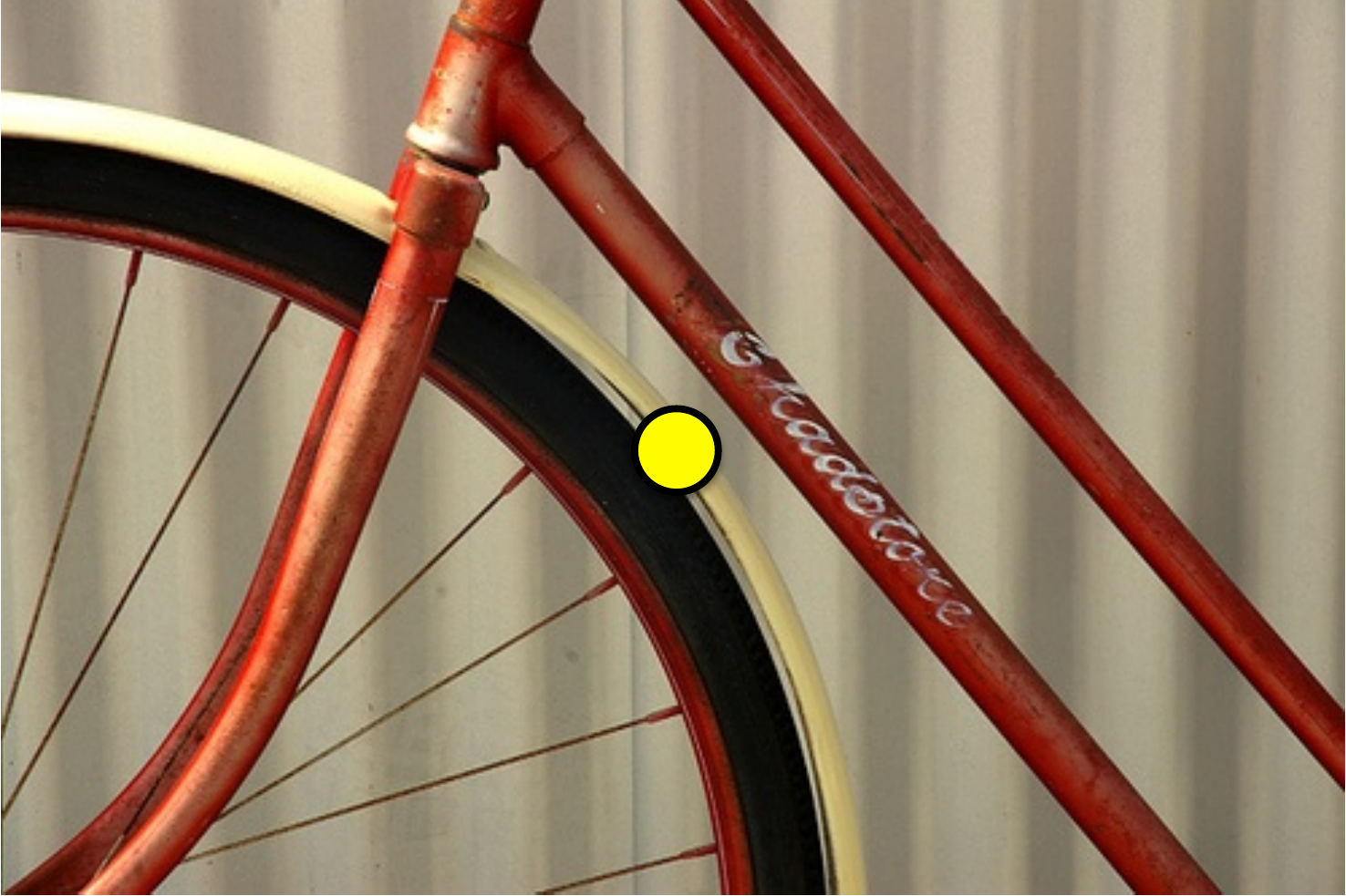} &
\textit{ } &
\includegraphics[width=0.32\linewidth]{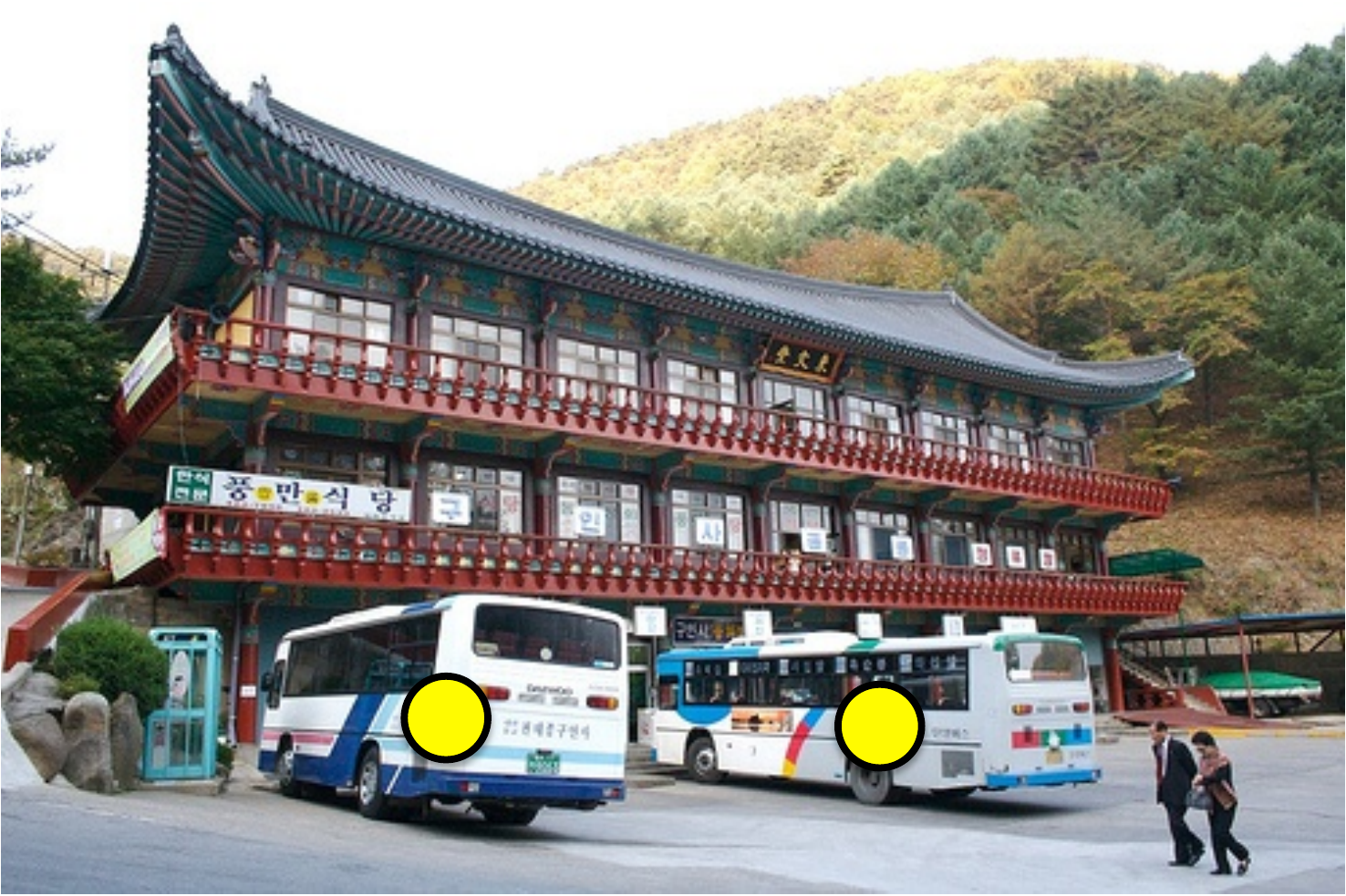}\\
%{\small{$(a)$}} &  & {\small $(b)$} &  & {\small $(c)$}
\end{tabular}
\caption{\small \textbf{Instruction Examples}: (left) the desired box center may not be on the object, (middle) if the object instance is truncated, click on the center of the visible part and (right) if multiple instances are present, click on the center of any one of them.}
\vspace{-.4cm}
\label{fig:mturkexamples}
\end{figure}

%In this section, we describe our method for collecting click annotations on images. The annotators are given an image and the name of a target object class and they are instructed to click once on the center of any instance of this class. 
We now describe the main components of our crowd-sourcing workflow, which is illustrated in Fig.~\ref{fig:mturkflow}. 

%The annotators read a set of instructions (Sec.~\ref{sec:mturkInstr}) and go through an interactive training stage (Sec.~\ref{sec:mturkQual}). 
%Annotators who successfully pass the training stage can proceed to the annotation stage (Sec.~\ref{sec:mturkTask}). In Sec.~\ref{sec:mturkResults}, we present the outcome of crowd-sourcing on PASCAL VOC 2007 and evaluate the quality of the clicks.

%paragraph of 1st draft
%In this section, we describe our method for collecting click annotations on images. The annotators are given an image and a target class and they are instructed to click once on the center of any instance of this class. The workflow of our Human Intelligence Task (HIT) \vitto{state we do AMT before saying HIT} is shown in fig.~\ref{fig:mturkflow}. Our HIT consists of two stages: the training stage and the actual annotation stage.During the training stage, the annotators should read a set of instructions (sec.~\ref{sec:mturkInstr}) about our task and pass a simple qualification test (sec.~\ref{sec:mturkQual}). Annotators who successfully pass the training stage can proceed to the actual annotation stage described in sec.~\ref{sec:mturkTask}. In section~\ref{sec:mturkResults} we present our crowd-sourcing results and evaluate the quality of the clicks.

\subsection{Instructions}
\label{sec:mturkInstr}

Our annotators are given an image and the name of the target class. Unlike~\cite{bearman16eccv} where annotators are asked to click anywhere on a target object, we want them to click on the center of an imaginary bounding box around the object (Fig.~\ref{fig:mturkexamples}).
This definition of center is crucial, as it provides a strong anchor point for the actual bounding box location.
However, humans have a tendency to click on the center of mass of the object, which gives a less precise
anchor point for the box location. We therefore carefully phrase our instructions as: ``imagine a perfectly tight rectangular box around the object and then click as close as possible to the center of this imaginary box''. For concave objects, the box center might even lie outside the object (Fig.~\ref{fig:mturkexamples}-left).

% of the bounding-box center 
% (fig.~\ref{fig:mturkexamples}(a)). For this reason, the task requires carefully worded instructions; we tell
% annotators to imagine a perfectly tight rectangular box around the object and then to click as close as 
% possible to the center of this imaginary box.

We also include explanations for special cases: If an object is truncated (i.e. only part of
it is visible), the annotator should click on the center of the visible part
(Fig.~\ref{fig:mturkexamples}-middle). If there are multiple instances of the target class, one should click on the center of only one of them (Fig.~\ref{fig:mturkexamples}-right).
% In order to illustrate our instructions, we provide the three examples images shown in fig~\ref{fig:mturkexamples}.

In order to let annotators know approximately how long the task will take, we suggest a time of 3s per click. This is an upper bound on the expected annotation time that we estimated from a small pilot study.

%paragraph of 1st draft
%\vitto{let's a smoother progression: from just click anywhere (bearman16eccv), to click in the center, to defining precisely what center we mean}
%\vitto{the whole subsection focuses on a specification of the generic task; in practice I believe we also tell them something basic, like 'click on the center of any instance of class X'. As your title is 'instructions', you should tell this basic fact, and only then delve into details of what does it mean 'center'; OR: change the title of the subsection}
%The goal of our method is to derive bounding-boxes from user clicks \vitto{really? Sounds like we directly produce BBs out of clicks; soften it: something like use clicks to guide a WSOL algorithm that tries to put a box on the object}. For this reason, we instruct the annotators carefully how they should estimate our desired center. Humans tend to think about the center of the mass of the object that may differ a lot from the center of a perfectly tight bounding-box around it (fig.~\ref{fig:mturkexamples}(a)). The right approach is to make them imagine a perfectly tight rectangular box around the object and then to click as close as possible to the center of this imaginary box. 
%In order to train the annotators on how they should respond correctly for our task on some particular cases, we provide three image examples shown in fig.~\ref{fig:mturkexamples}. 
% (a) center out of object
% (b) multiple instances
% (c) truncated instances

\subsection{Annotator training}
\label{sec:mturkQual}

\begin{figure}[t]
\vspace{-.6cm}
\center
\includegraphics[width=1\linewidth]{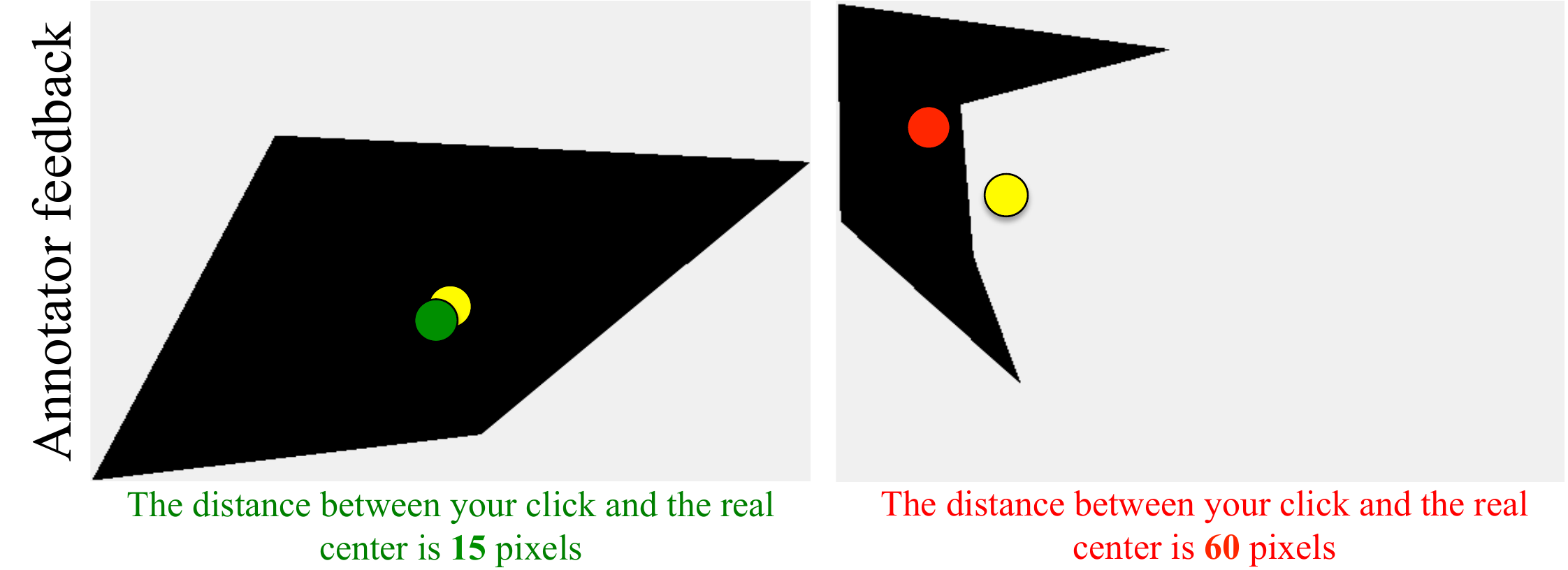}
\caption{\small Examples that the annotators receive as feedback. For each example, we provide the real center of the polygon (yellow dot), their click (green or red dot) and the Euclidean distance between the two.}
\label{fig:mturkFeedback}
\vspace{-.4cm}
\end{figure}

After reading the instructions, the annotators go through the training stage. They complete a simple qualification test, at the end of which we provide detailed feedback on how well they performed. Annotators who successfully pass this test can proceed to the annotation stage. In case of failure, annotators can repeat the test until they succeed.

%\jasper{Next sentence is Too early. Should be in conclusion of whole section.}
%The combination of providing rich feedback and allowing annotators to repeat test and feedback makes this a highly effective, interactive training stage.

\mypar{Qualification test.}
%Having a qualification test is a good mechanism for enhancing the quality of the crowd-sourcing data and filtering out bad annotators and spammers~\cite{andriluka14cvpr,endres10cvprw,Johnson11cvpr,krause2013iccvw}. Some annotators do not pay attention to the task instructions and others do not even read them. Qualification tests have been successfully used for collecting image labels, drawing object bounding-boxes, and obtaining segmentations for some of the most popular computer vision datasets (e.g.,~MS COCO~\cite{lin14eccv} and Imagenet~\cite{russakovsky15ijcv,su12aaai}). 
%
Qualification tests have been successfully used for enhancing the quality of the crowd-sourced data
and filtering out bad annotators and
spammers~\cite{andriluka14cvpr,endres10cvprw,Johnson11cvpr,krause2013iccvw,russakovsky15ijcv,su12aaai}.
This happens because some annotators pay little to no attention to the task instructions. 

During a qualification test, the annotator is asked to respond on some questions for which the answers are known. This typically requires experts to annotate a batch of examples (in our case draw object bounding boxes). 
Instead, we use an annotation-free qualification test in which the annotators need to click on the center of $20$ synthetically generated polygons, like the ones in Fig.~\ref{fig:mturkflow}.
Using synthetic polygons allows us to generate an arbitrarily large set of qualification questions with zero human annotation cost.
Additionally, annotators cannot overfit to qualification questions or cheat by sharing answers, which 
is possible when the number of qualification questions is small.

% The qualification test should mimic our main task on clicking on the center of the objects in natural images. The annotators should not be able to pass the test successfully if they do not pay attention to our instruction set or do not fully understand the task. They are shown $20$ different polygons sequentially and are asked to click on their center. The position, the size, the shape and the number of angles of each polygon are randomly generated for each example. The polygons can be either convex or concave. The generated images that contain the polygons have a constant size of $500\times333$ pixels. The mean area of the generated polygons is about 50k pixels which corresponds to about one third of the image. \jasper{Too much repetition. Also, pixel sizes are boring. The figure with feedback is detailed enough.

\mypar{Why polygons?}
We use polygons instead of axis-aligned rectangles in order to train the annotators on the difference between the center of mass of an object and the center of the imaginary box enclosing the object. Moreover, polygons provide a more realistic level of difficulty for the qualification test. Finding the center of an axis-aligned rectangle is trivial, whereas finding the center of a polygon is analogous to finding the center of a real object.
And yet, polygons are abstractions of real objects, thus reducing the cognitive load on the annotators, potentially making the training stage more efficient.

\mypar{Feedback.}
After the annotators finish the qualification test, they receive a feedback page with all polygon examples they annotated (Fig.~\ref{fig:mturkFeedback}). For each polygon, we display (a) the position of the real center, (b) the position of the annotator's click, and (c) the Euclidean distance in pixels between the two (error distance).

\mypar{Success or failure.}
The annotator needs to click close to the real centers of the polygons in order to pass the test.
%\dimpp{ We set this threshold distance at 10\% of the square root of the polygon area. Examples of polygons with the acceptance area for clicks around the real center can be seen in fig.~\ref{fig:mturkFeedback}. The mean square root of the areas is about 200 pixels. An annotator therefore passes the test if the mean distance over all the examples is less than 20 pixels.}
The exact criterion to pass the test is to have an error distance below $20$ pixels, on average over all polygons in the test.

The annotators that pass the qualification test are flagged as \textit{qualified annotators} and can proceed to the main annotation task where they work on real images. A qualified annotator never has to retake the qualification test.
In case of failure, annotators are allowed to repeat the test as many times as they want until they pass it successfully. 

The combination of providing rich feedback and allowing annotators to repeat the test results in an interactive and highly effective training stage.

%After the worker finishes the qualification test, he receives a feedback page with all polygon examples he \vitto{sexist} annotated. For each polygon example, we provide \vitto{do you mean we provide this to the annotator/worker as a feedback?} the position of the real center, the position of the worker's click and the euclidean distance in pixels between the two. The workers pass the test if the mean distance over all the examples is less than 20 pixels \vitto{meaningless threshold: earlier on, when you describe the generation process, tell about your image sizes and average polygon sizes}. In case of failure, the annotator has the option to repeat the test as many times as he wants until he passes it successfully \vitto{we should state that the combination of providing detailed feedback and allowing for repeating the test on a new set of polygons effectively makes this protocol a highly effective, interactive training stage}. The annotators that pass the qualification test are flagged as \textit{qualified annotators} and can proceed on the main task where they work on real images. A qualified annotator can skip the qualification test if he accepts further HITs. This means that each annotator should pass the qualification test only once. 

%\dimpp{Do we need a figure for the feedback?}
%\vitto{if you really show it to the annotator, yes, it would be cool, also plays the role of support figure when explaining how we estimate error models later on}

\subsection{Annotating images}
\label{sec:mturkTask}

\begin{figure}[t]
\vspace{-.4cm}
\center
\includegraphics[width=1\linewidth]{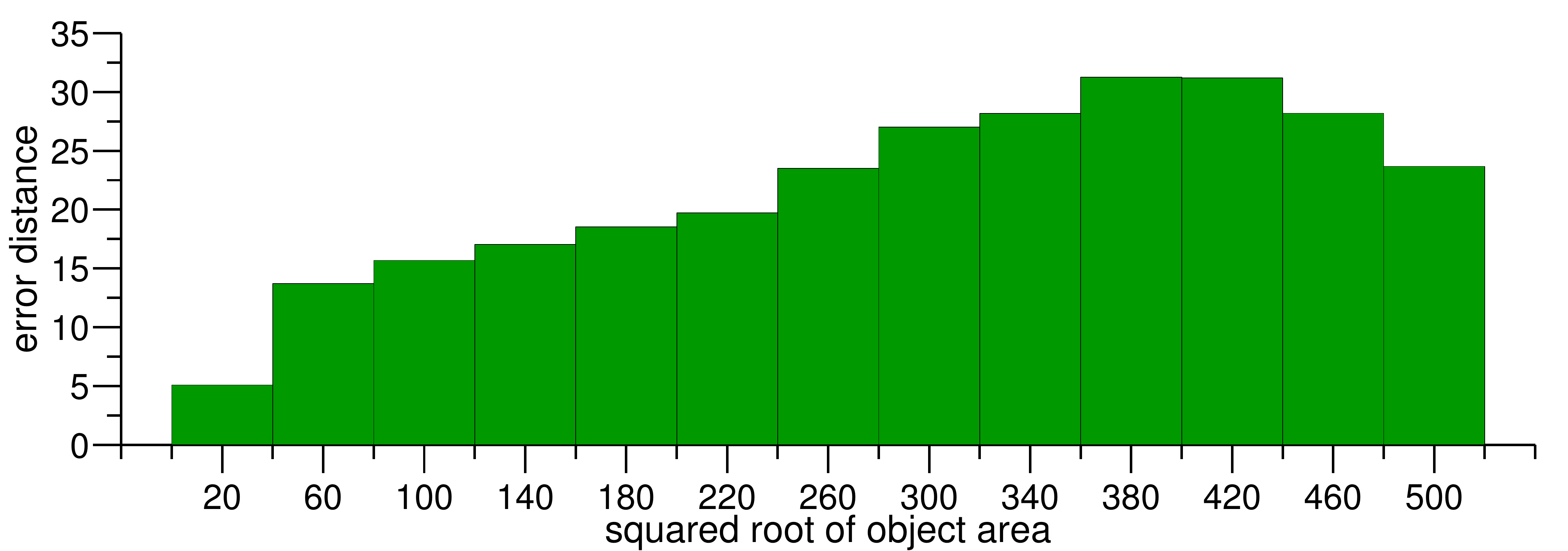}
\caption{\small The error distance of the annotators as a function of the square root of the object area.}
\vspace{-.4cm}
\label{fig:mturkErrorArea}
\end{figure}

In the annotation stage, annotators are presented small batches of $20$ consecutive images to annotate. 
For increased efficiency, our batches consist of a single object class. Thanks to this, annotators do not have to re-read the class name for every image, and can keep their mind focused on their prior knowledge about the class to find it rapidly in the image~\cite{Torralba:ea:06}. More generally, it avoids task-switching which is well-known to increase response time and decrease accuracy \cite{Rubinstein:ea:01}.
%\vitto{well, good point but there is more to it imho, beyond simply not having to read the name of a new class evry time; they can also focus they attention to one class for many images, so they do not need to 'switch expected shapes' in their head, accelerating visual search; it would be ideal to have a cogsci citation for this effect: Frank to the rescue?}. 

%As mentioned above, annotators that successfully pass the qualification test \vitto{qualification vs training; stage vs phase; task vs test} once are qualified to proceed to the main work of the HIT. In this stage they are asked to annotate a small batch of $20$ images.The target class for all the images within a batch is the same. As mentioned above (\dimpp{where?}) we assume access to image-level labels, so we know that at least one instance of the target class in presenting in an image \vitto{standard WSOL setting}. Using the same target class for the whole batch speeds up a lot the response time of the annotators as they know the target class even before looking at the next image \vitto{well, good point but there is more to it imho, beyond simply not having to read the name of a new class evry time; they can also focus they attention to one class for many images, so they do not need to 'switch expected shapes' in their head, accellerating visual search; it would be ideal to have a cogsci citation for this effect: Frank to the rescue?}.

\mypar{Quality control.}

%\cite{sorokin08cvprw,welinder10nips}
%\cite{bearman16eccv,lin14eccv,russakovsky15ijcv,su12aaai}
%\cite{kovashka15ijcv} \cite{russell08ijcv} \cite{vondrick13ijcv}

Quality control is a common process when crowd-sourcing image annotations~\cite{bearman16eccv,kovashka15ijcv,lin14eccv,russakovsky15ijcv,russell08ijcv,sorokin08cvprw,su12aaai,vondrick13ijcv,welinder10nips}.
We control the quality of click annotation by hiding two evaluation images for which we have ground-truth bounding boxes inside a 20-image batch, and monitor the annotator's accuracy on them (golden questions). Annotators that fail to achieve an accuracy above the threshold set in the qualification test are not able to submit the task. We do not do any post-processing of the submitted data.

We point out that we use extremely few different golden questions, and add them repeatedly to many batches. On PASCAL VOC 2007, we used only $40$, which amounts to $0.5\%$ of the dataset. This is a negligible overhead.
%\vitto{danger, by admitting to the quality control with golden questions, we are saying we use GT BBs; we need to state how many in total, to calm down the reader; now it sounds like 10\% of the whole dataset! Hopefully it's less, as some GQ are repeated across batches?} \frank{Exactly, we need to quantify this.} \dimpp{I always use the same 2GQ per class. So in total i used 40 boxes! that's only 0.5\% of the whole dataset}

%Quality control during the main task is also important \dimpp{hmmm REF??} \vitto{we can state is a common thing to do in crowdsourcing for image annotations, and cite tons of papers that say they do it}. We control the quality of the click annotation by hiding 2 evaluation images inside a 20-image batch and monitoring the annotator's accuracy on these 2 images \vitto{this golden questions type of approach is cool, but it also costs some manually drawns BBs; you just argued in sec. 3.2 that the qualification test we propose avoid this additional cost; somehow we are in a tension on this point; we should discuss in person}. If the annotator fails to keep an accuracy similar to the one of the qualification test, we do not let him submit the results of the task. Otherwise, he simply submits the HIT and is free to continue working on further HITs. We do not do any post-processing rejection of the submitting data and we accept all of them.

\subsection{Data collection}
\label{sec:mturkResults}

We implemented our annotation scheme on Amazon Mechanical Turk (AMT) and we collected click annotations for all 20 classes of the whole trainval set of PASCAL VOC 2007~\cite{everingham10ijcv}. Each image was annotated with a click by two different annotators for each class present in the image. This results in 14,612 clicks in total for the 5,011 trainval images.

%We implemented our system on the Amazon Mechanical Turk (AMT) crowdsourcing platform \vitto{I believe this should be said earlier, as you use the AMT-specific tech term 'HIT'} and we collected click annotations for the whole trainval set of the PASCAL VOC 2007 dataset~\cite{everingham10ijcv}. Each image was annotated with a click by $2$ different AMT workers for each one of the 20 PASCAL classes presented in this image \vitto{weird, unclear, it sounds like all 20 classes are present in each image; rewrite}. We paid the workers $\$0.10$ to annotate a batch of 20 images. This results in a standard wage of 6\$ per hour for a generous estimation of 3 seconds per image \vitto{not so generous as we gave no idea how long it takes so far; maybe we can state that we knew it takes at most that time given preliminary in-house studies; OR: we delay talking about money until much later; anyway money does not matter too much; it's time that matters in this kind of paper}. 

%The total cost for annotating the whole trainval set of PASVAL VOC 2007 with 2 click annotations per class was $\$75.4$ (or $\$37.7$ for 1 click annotation).

\mypar{Annotation time.}

\begin{table}[t]
\vspace{-.3cm}
\centering
\small
\begin{tabular}{|c|c||c|} \hline
Qualification test & Quality control &  Error distance\\
\hline
No & No & 43.8 \\
images & No & 29.4 \\
polygons & No & 29.3 \\
polygons & Yes & 21.2 \\
\hline
\end{tabular}
\caption{\small The influence of the two main elements of our crowd-sourcing protocol on click accuracy.}
\vspace{-.4cm}
\label{tab:mturkErrorTable}
\end{table}

%\vitto{the order of sec. 3.4 is not ideal; you discuss time above already above, and now you actually measure it, weird}

During the annotation stage we measure the annotator's response time from the moment the image appears %(onset of the image) 
until they click. The mean response time was $1.87$s. % while the median time was $1.54$s.
%(unlike~\cite{bearman16eccv,su12aaai}, we use the mean time as this leads to the real time to annotate the whole dataset 
%\vitto{we use where? So far we reported both; maybe this comment belongs in the result section?}). \frank{I think we should just get rid of the mean/median disucssion. We report both, so we don't need to justify what we're doing.}
This indicates that the task can be performed very efficiently by annotators. Note that we are able to annotate the whole PASCAL VOC 2007 trainval set with one click per object class per image in only 3.8 hours. 

%This is a $22\times$ speedup comparing to the efficient bounding-box annotation protocol of~\cite{su12aaai}. \vitto{unclear, dangerous; su12aaai does not do PASCAL VOC, and does not report total times; better to keep the comparison to the time taken for 1 instance, so: 42s/88s for su12aaai and 1.87s for us; only instance to instance comparisons possible!} \vitto{and even then, they annotated BBs, we annote points, so really we cannot talk about speedups at this point in the paper; we can say that later, after we derive BBs from these points}

Interestingly, the response time we measured is comparable to image-level annotation time (1.5s in \cite{krishna16chi}) indicating that most of the time is spent on the visual search to find the object and not on clicking on it. Also, our requirement to click on the center of the object does not slow down the annotators: our response time is comparable to the time reported in~\cite{bearman16eccv} for click-anywhere annotations.
%: 2.4s in~\cite{bearman16eccv}. \vitto{this is weird, and probably due to how well we trained them? We should either state it does not slow down and that's it, or, if we claim is faster, then attribute it hypothetically to something. Frank?} \frank{This is a small difference, and there are lots of differences between us and \cite{bearman16eccv} belides center/anywhere click, so we can't really claims based on this difference. We would have to run an experimet with anywhere-click in our setup. I have thus removed this claim.}

We examined the response time as a function of the area of the target object and we observed an interesting phenomenon. Response time does not increase when the object becomes smaller, ranging from 1.7s for very small objects to 2.2s for object as big as the whole image. We hypothesize that while small objects are more difficult to find, estimating their center is easier than for large objects.
%\vitto{maybe we should claim a little less: just say time does not increase when the object becomes smaller, and that's it; claiming the reverse fx is weird, especially without backing it up with more evidence; Frank?} \frank{I've toned this down. And made clear that we're just speculating. The important thing is that reponse time doesn't increase with size, as it would for search, for example.}

%\dimpp{It would be great here if we can find a similar analysis on other tasks (image-level labels, draw boxes, etc...)}

%\vitto{define relative area}
%In fig.~\ref{fig:mturkRT}, we examine the human response time as a function of the relative area of the object. Interestingly, we can observe that the response time is not higher when the object becomes smaller. Actually the response time is slightly increasing for big objects as estimating the center there is a more difficult task. The time ranges from 1.7s for small objects (0.1 relative area) to 2.2s for whole image objects. 
%So, we expect that the response time per image will go down if one uses our crowd-sourcing method to annotate more difficult datasets.
%\vitto{perhaps true in some absolute mathematical sense, but the variations in response time as a fct of area are small, relative to the mean time; let's not claim this; it also opens a trap: we do COCO, which has smaller objects than VOC, but we do not do it with real annotators!}

%\dimpp{don't like the figure: I tried to explain it in the text. We can easily remove it}
%\vitto{the figure does not bring much, but in conjuction to the paragraph it's quite interesting}

\mypar{Error analysis.}
We evaluate the accuracy of the collected clicks by measuring their distance from the true centers of the ground-truth object bounding boxes. In Fig.~\ref{fig:mturkErrorArea} we show this error distance as a function of the square root of the object area.
As expected, the error distance in pixels increases as the object area increases. However, it slightly drops as the object occupies the whole image. This is likely because such images have truncated instances, which means the annotator needs to click in the center of the image rather than the center of the object, an easier task. 
In general, the error distances are quite low: 19.5 pixels on average with a median of 13.1 pixels (the images are 300x500 on average).
%\vitto{what about the average error divided by the sqrt area? This is interesting!} \dimpp{you mean object area right? Do you want me to change all numbers of this section??} \dimpp{mean:0.15, median:0.09}

%The annotators were highly accurate at the task. with a mean error distance of 19.5 pixels (the median error distance was only 13.1 pixels).
%\vitto{19.5 is not so meaningful: it depends on how big objcts are on average; I'd like to see errors as a fct of sqrt obj area, that's interesting!}
%\vitto{19.5 is very very close to the threshold of the qualification test; makes that test sound fragile!}

Next, we want to understand the influence of using a qualification test, using quality control, and using polygons or real examples during the qualification test. Therefore we conducted a series of smaller-scale crowd-sourcing experiments on 400 images of PASCAL VOC 2007 trainval.
As Tab.~\ref{tab:mturkErrorTable} shows, using a qualification test reduces average error substantially, from 43.8 to 29.4 pixels. Interestingly, using polygons instead of real examples does not influence the error at all, demonstrating that our proposed qualification test is well-suited to train annotators.
Quality control, hiding two evaluation images inside the task of annotating images, brings the error further down to 21.2 pixels (on the full dataset we measure 19.5 pixels error).
%We conclude that both the qualification test and quality control improve annotation quality \vitto{tricky: this means it's crucial to have those extra manually annotated qual images; maybe we can say something weaker here} \frank{Toned down claim.}
%
Finally, we note that all four variants in Tab.~\ref{tab:mturkErrorTable} resulted in similar annotation time. Hence qualification tests or quality control has no significant influence on the speed of the annotators.

\mypar{Cost.}

We paid annotators $\$0.10$ to annotate a batch of 20 images. Based on their mean response time this results in a wage of about $\$9$ per hour. The total cost for annotating the whole trainval set of PASCAL VOC 2007 with two click annotations was $\$75.40$ (or $\$37.70$ for one click annotation).

\section{Incorporating clicks into WSOL}

\begin{figure}[t]
\vspace{-.1cm}
\center
\begin{tabular}{c@{}c@{}c@{}c@{}c@{}}
\includegraphics[width=0.32\linewidth]{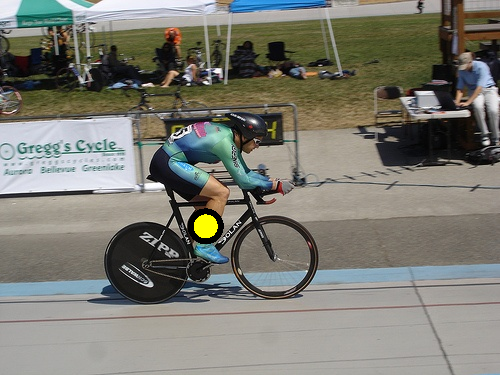}&
\textit{ } &
\includegraphics[width=0.32\linewidth]{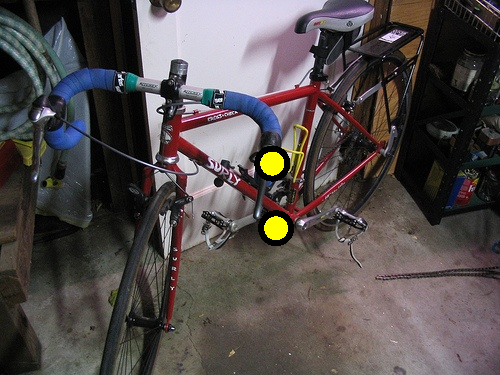}&
\textit{ } &
\includegraphics[width=0.32\linewidth]{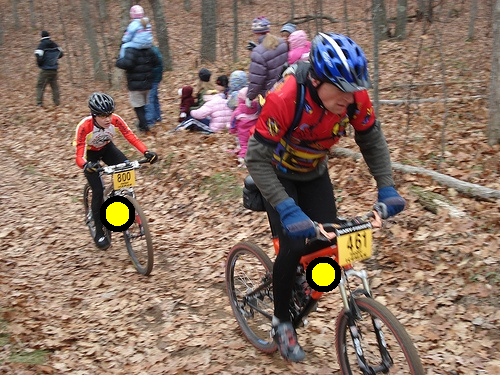} \\
\includegraphics[width=0.32\linewidth]{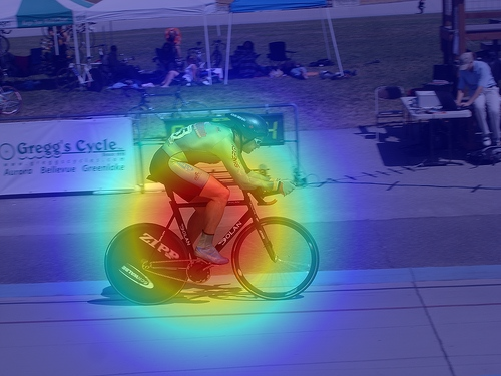}&
\textit{ } &
\includegraphics[width=0.32\linewidth]{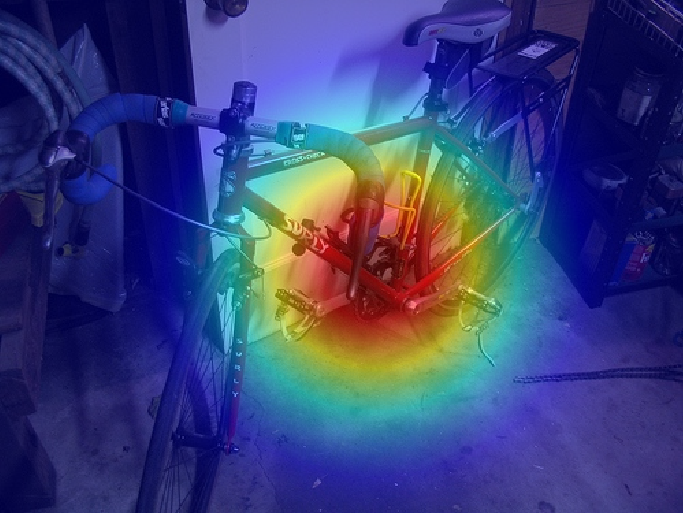}&
\textit{ } &
\includegraphics[width=0.32\linewidth]{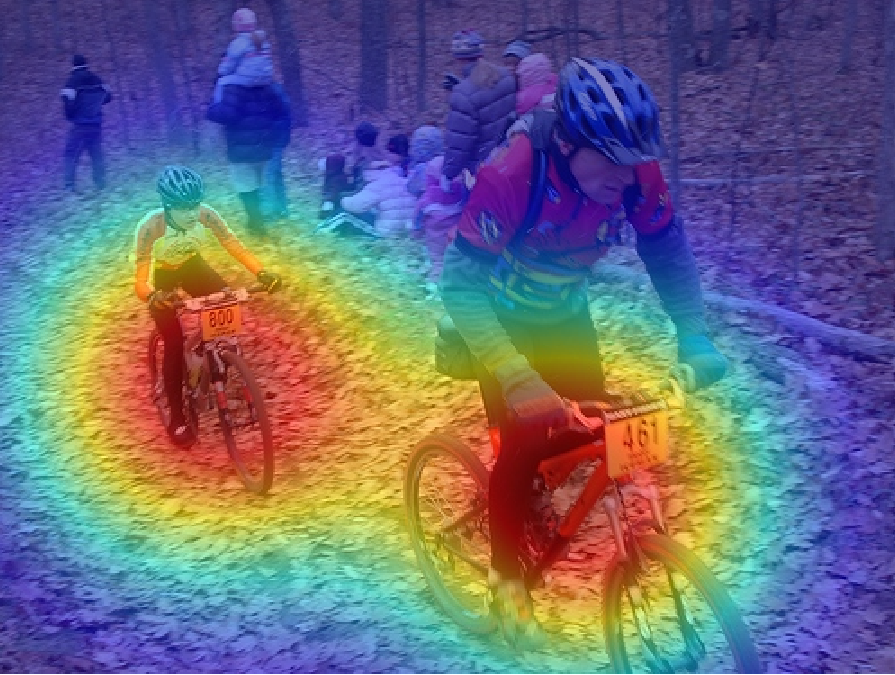} \\
%{\small{$(a)$}} &  & {\small $(b)$}
\end{tabular}
\caption{\small \textbf{Box center score $S_{bc}$} on bicycle examples.
(left): One-click annotation. (middle): Two-click annotation on the same instance. (right): Two-click annotation on different instances. The values of each pixel in the heatmaps give the $S_{bc}$ of an object proposal centered at that pixel.}
\vspace{-.4cm}
\label{fig:methodSBCexample}
\end{figure}

We now present how we incorporate our click supervision into a reference Multiple Instance Learning (MIL) framework, which is typically used in weakly supervised object detection (WSOL). All explanations in this section consider working with one object class at a time, as we treat them essentially independently.

% We present here our method that incorporates click supervision inside a WSOL framework. We first describe a reference MIL framework that serves as our baseline~\ref{sec:methodMIL}. In sec.~\ref{sec:method1click} we show how to extend this framework: we use one-click annotations on the center of the object to estimate object location. In sec.~\ref{sec:method2click} we show how to estimate object location and area when center clicks from two annotators are available.

%We present here our method for localizing objects in an image under click supervision \vitto{almost right, but not it sounds like collomosse's variant of grabcut; instead we do WSOL under click supervision, so: multiple images of a class, pos and neg}. 
%We first describe a standard MIL framework \vitto{standard? I'd like to say 'reference', because it's not super basic nor super standard, it's a 'gold standard', it's good! It's got objectness and multi-folding, it's got proper initialization, lots of engineering, we need to position our baseline/reference positively} that serves as our baseline~\ref{sec:methodMIL}. In section~\ref{sec:method1click} we show how to extend this framework by using one-click annotations on the center of the object for estimating the object location. In section~\ref{sec:method2click} we show how to estimate the object location and area by having 2 annotators click once on the center of an object.

\subsection{Reference Multiple Instance Learning (MIL)}
\label{sec:methodMIL}

The input to MIL is a training set with positive images, which contain the target class, and negative images, which do not.
We represent each image as a bag of object proposals extracted using Edge-Boxes~\cite{dollar14eccv}. Following~\cite{girshick14cvpr,cinbis15pami,bilen14bmvc,bilen15cvpr,song14icml,wang15tip}, we describe each object proposals with a 4096-dimensional feature vector using the Caffe implementation~\cite{jia13caffe} of the AlexNet CNN~\cite{krizhevsky12nips}. We pre-trained the CNN on the ILSVRC~\cite{russakovsky15ijcv} dataset using only image-level labels (no bounding box annotations).

A negative image contains only negative proposals, while a positive image contains at least one positive proposal, mixed in with a majority of negative ones. The goal is to find the true positive proposals from which to learn an appearance model for the object class. We iteratively build an SVM appearance model $\mathcal{A}$ by alternating between two steps:

(I) {\em re-localization}: in each positive image, we select the proposal with the highest score given by the current appearance model $\mathcal{A}$.

(II) {\em re-training}: we re-train the SVM using the current selection of proposals from the positive images, and all proposals from negative images.

As initialization, in the first iteration we train the classifier using complete images as positive training examples~\cite{cinbis14cvpr,cinbis15pami,pandey11iccv,russakovsky12eccv,nguyen09iccv,kim2009nips}.

\mypar{Refinements.}
In order to obtain a competitive baseline, we apply two refinements to the standard MIL framework. First, we use multi-folding~\cite{cinbis15pami}, which helps escaping local optima. Second, we combine the score given by the appearance model $\mathcal{A}$ with a general measure of ``objectness''~\cite{alexe10cvpr} $\mathcal{O}$, which measures how likely it is for a proposal to tightly enclose an {\em object} of any class (e.g. bird, car, sheep), as opposed to background (e.g. sky, water, grass). Objectness was used in WSOL before, to steer the localization process towards objects and away from background~\cite{cinbis15pami,deselaers10eccv,guillaumin12cvpr,prest12cvpr,shapovalova12eccv,siva11iccv,shi12bmvc,tang14cvpr,wang14eccv-cosegmentation}. In this paper we use the recent objectness measure of~\cite{dollar14eccv}.

%\vitto{I do not think we need the following} \dimpp{we define $S_{ap}$ that we use later}
Formally, at step (I) we linearly combine the scores $\mathcal{A}$ and $\mathcal{O}$ under the assumption of equal weights. The score of each proposal $p$ is given by $S_{ap}(p)=\frac{1}{2} \cdot \mathcal{A}(p)+ \frac{1}{2} \cdot \mathcal{O}(p)$.

\mypar{Deep MIL.}
After MIL converges (typically within 10 iterations), we perform two additional iterations where during the step (II) we deeply re-train the whole CNN network, instead of just an SVM on top of a fixed feature representation. During these iterations we use Fast RCNN~\cite{girshick15iccv} as the appearance model $\mathcal{A}$.

\subsection{One-click supervision}
\label{sec:method1click}

%\vitto{slash-para this section, so we can highlight motivation, model, and two-fold usage: during re-localization and during initialization}

\paragraph{Motivation.}
Click annotations on object centers derived using our crowdsourcing method of Sec.~\ref{sec:mturk} provide a powerful cue about object position. In this section, we improve the reference MIL framework by using the position of one single click $c$ in each image of the target class.
%re-localization step (I) of the reference MIL.

% As our annotators were carefully instructed to click on the center of an imaginary box tightly enclosing the target object, one could simply pick as the object localization the object proposal whose center is the nearest to the annotator's click. However, this approach would probably fail since the positions of the annotator's clicks do not perfectly match the center of the object (sec.~\ref{sec:mturkResults}) 

\mypar{Box center score $S_{bc}$.}
Intuitively, simply selecting the object proposal whose center is closest to the click would fail since annotators are not perfectly accurate. Instead, we introduce a score function $S_{bc}$, which represents the likelihood of a proposal $p$ covering the object according to its center point $c_p$ and the click $c$
\vspace{-.2cm}
\begin{equation}
\label{eq:sbc}
S_{bc}(p;c,\sigma_{bc}) = \mathrm{e} ^ {-{\frac{{ \lVert {c_{p}-c} \rVert }^2}{2\sigma_{bc}^2}}}
\end{equation}
\noindent where $\lVert {c_{p}-c} \rVert$ indicates the Euclidean distance in pixels between $c$ and $c_p$.
The standard deviation $\sigma_{bc}$ controls how quickly the $S_{bc}$ decreases as $c_p$ gets farther from $c$ (Fig.~\ref{fig:methodSBCexample}).

\mypar{Use in re-localization.}
We use the box center cue $S_{bc}$ in the re-localization step~(I) of MIL (Sec.~\ref{sec:methodMIL}).
Instead of selecting the proposal with the highest score according to the score function $S_{ap}$ alone, we combine it with $S_{bc}$ with a product: $S_{ap}(p) \cdot S_{bc}(p;c,\sigma_{bc})$.
In Sec.~\ref{sec:resultsVOC} we show that this results in improved re-localization, which in turn leads to better appearance models in the next re-training iteration, and ultimately improves the final MIL outcome.
%Also, using this cue deeply inside the MIL from the first iteration improves the appearance model $\mathcal{A}$ since it is now trained from a set $\mathcal{P}$ which typically contains better positive training examples.

\mypar{Use in initialization.}
We also use the click position to improve the MIL initialization.
Instead of initializing the positive training samples from the complete images, we now construct windows centered on the click while at the same time having maximum size without exceeding the image borders. This greatly improves MIL initialization, especially in cases where the position of the click is close to the image borders.
%\dimpp{removed reference to results sec as we don't show that anymore.}
%(Sec.~\ref{sec:resultsVOC}).

%\dimpp{Should I mention how I learn sigmaBC here? I learn all parameters from the polygons! I want to have a subsection at the end of the section, explaining how I learn all these parameters (sigmaBC, sigmaBA, meanBA, maxD)}
%\vitto{that's a great point of our work: it's healthy machine-learning-wise, it's new in the WSOL area, and most importantly it brings attention to one of our novel elements (the polygons).}
%\vitto{I am not sure whether it's better to have a slash-para here and another one in sec. 4.3 about this hyper-param training, or a single subsec. 4.4. Let's see...}

\subsection{Two-click supervision}
\label{sec:method2click}

\begin{figure}[t]
\vspace{-.4cm}
\center
\begin{tabular}{c@{}c@{}c@{}}
%\includegraphics[width=0.49\linewidth]{BA_small.png}&
%\textit{ } &
%\includegraphics[width=0.49\linewidth]{BA_small_Windows.png}\\
\includegraphics[width=0.49\linewidth]{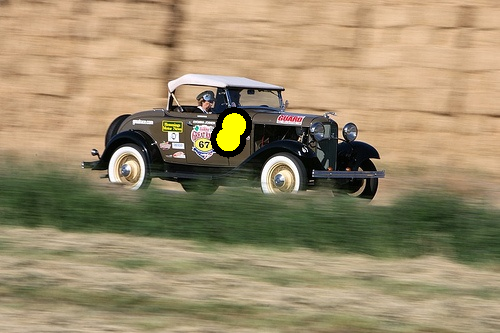}&
\textit{ } &
\includegraphics[width=0.49\linewidth]{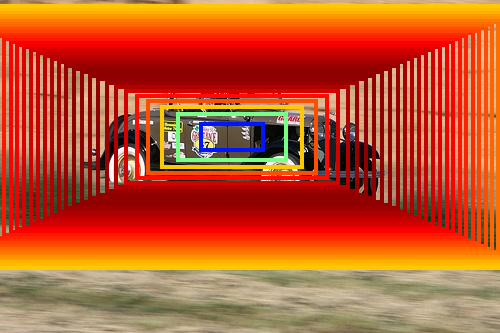}\\
%\includegraphics[width=0.49\linewidth]{BA_big.png}&
%\textit{ } &
%\includegraphics[width=0.49\linewidth]{BA_big_Windows.png}\\
%{\small{$(a)$}} &  & {\small $(b)$}
\end{tabular}
\caption{\small \textbf{Box area score $S_{ba}$}. All windows used here have fixed aspect ratio and are centered on the center of the object.}
\vspace{-.4cm}
\label{fig:methodSAexamples}
\end{figure}

\paragraph{Motivation.}
% In sec.~\ref{sec:mturkResults}, we collected two clicks $c_1$ and $c_2$ from two different annotators on the center of a target object in an image. Here, we use both of them to extend the MIL framework. 
%
%In this section, we extend the MIL framework by having two different annotators clicks $c_1$ and $c_2$ on the center of a target object in an image. \vitto{not quite, in the flow of the paper, we did this already in section 3.4; we need to rephrase somehow}.
%
While using two annotator clicks doubles the total annotation time compared to one click, it allows us to  estimate the object center even more accurately. Moreover, we can estimate the object area based on the distance between the two clicks.

%we define a \textit{box area} score $S_{ba}$ that represents the likelihood that a proposal $p$ covers the object according to the size estimate $\mu_{ba}$. We derive this size estimate by looking solely at the distance between the two clicks \vitto{reverse this explanation: say the two clicks enable us to estimate the object area/size (do not tell how, point to sec. 4.3.2); and then say we derive a new score using this area/size estimate}. 

\mypar{Box center score $S_{bc}$.}

%\vitto{reverse explanation; what matters most in this subsubsection is that two clicks enable us to estimate the object center more accurately; explain that without worrying about exceptions in the case of two instances, that's a corner case, it's not the core of the argument; after that explain your multi instance in one image case; after that tell that the actual box center score is as in eq (1) , just with another click center defined specially}

By averaging the positions of two clicks we can estimate the object center more accurately.
We simply replace $c$ in Eq.~\eqref{eq:sbc} with the average of the two clicks $c_1$ and $c_2$.

% Assuming that both clicks land on the same object, we create one single mode for the $S_{bc}$ score centered on the mean position of the two clicks. Taking the mean position improves the estimation of the object center (fig.~\ref{fig:methodSBCexample}(Middle)). The $S_{bc}$ score is defined in a similar way as in eq.~\ref{eq:sbc}, but now $c$ corresponds to the mean position of the two clicks $c_1$ and $c_2$.

However, in images containing multiple instances of the target class, the two annotators might click on different instances (Fig.~\ref{fig:methodSBCexample}, right). To address this, we introduce a distance threshold $d_{max}$ beyond which the clicks are considered to target different instances. In that case, we keep both clicks and use them both in Eq.~\eqref{eq:sbc}.
Formally, if $\lVert {c_1-c_2} \rVert > d_{max}$, then for each proposal $p$ we use the nearest of the two click to its center $c_p$.

\mypar{Box area score $S_{ba}$.}

There is a clear correlation between the area of the object and the click's error distance (Fig.~\ref{fig:mturkErrorArea}).
As errors made by two annotators are independent, the distance between their two clicks increases as the object area increases (on average). Therefore we estimate the object area based on the distance between the two clicks $c_{1}$ and $c_{2}$.

\begin{figure}[t]
\vspace{-.4cm}
\center
\begin{tabular}{c@{}c@{}c@{}}
\includegraphics[width=0.49\linewidth]{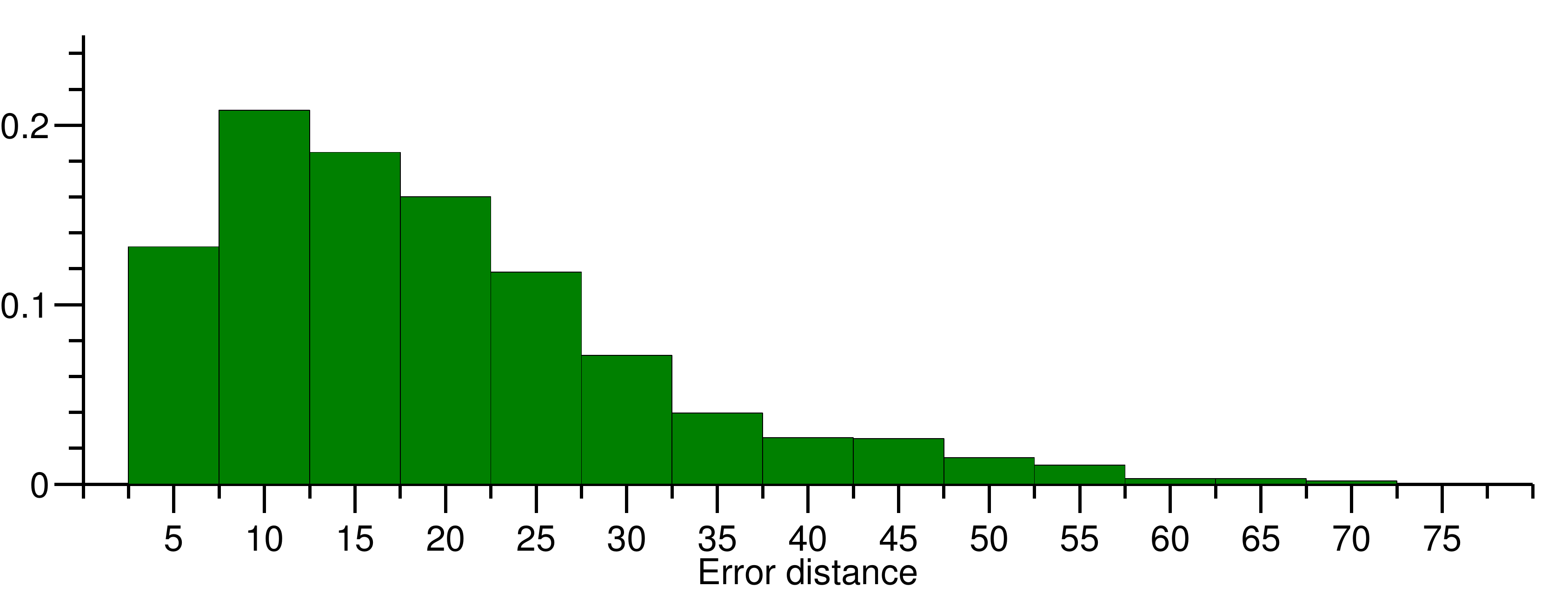}&
\textit{ } &
\includegraphics[width=0.49\linewidth]{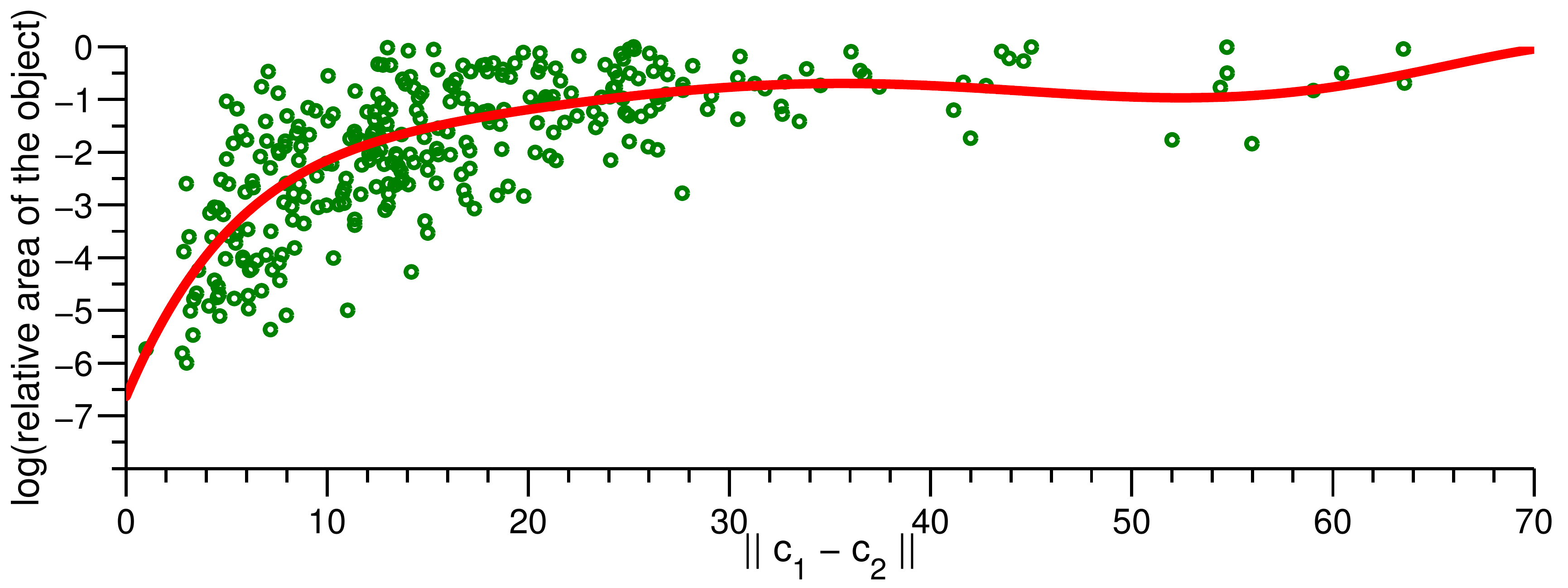}\\
%{\small{$(a)$}} &  & {\small $(b)$}
\end{tabular}
\caption{\small (left) The distribution of errors that the annotators made during our qualification test. (right) The relative area of the synthetic polygons (log scale) as a function of the distance between two clicks. The red line shows the regressed function $\mu$.}
\vspace{-.4cm}
\label{fig:methodPolyErrors}
\end{figure}

\begin{figure*}[t]
\vspace{-.5cm}
\center
\includegraphics[width=1\linewidth]{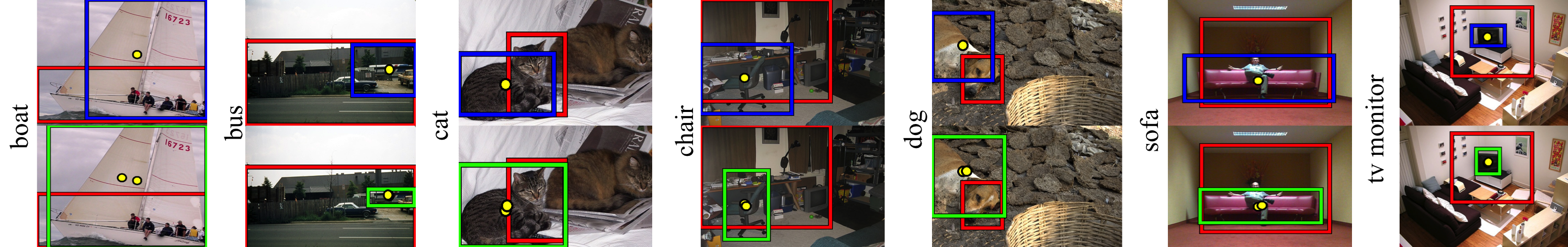}
\vspace{-.3cm}
\caption{\small Examples of objects localized on the trainval set of PASCAL VOC 2007 using our one-click (blue) and two-click (green) supervision models. For each example, we also show the localization produced by the reference MIL (red).}
\vspace{-.4cm}
\label{fig:resultsQualExamples}
\end{figure*}

% As already analyzed in sec~\ref{sec:mturkResults}, the annotators find it more difficult to accurately estimate the object center when the object in the image is big. More importantly, there is a clear correlation between the error distance and the area of the object (fig.~\ref{fig:mturkErrorArea}). Assuming that annotator error is random, this means that the distance between the two clicks increases when the area of the object increases. This can be observed in fig.~\ref{fig:methodSAexamples}(a). When the object is small (top example) the two clicks land almost on the same pixel. The distance between the two gradually increases when moving to medium-sized and large objects. 

% We can therefore estimate the object area based on the distance between the to clicks. The area estimate can then be used to refine the re-localization step of MIL. Once more, we do not naively pick the object proposal whose area is the nearest to the estimate, mostly because the area estimate cannot be perfect.

%\vitto{too complex flow; let's first say how we estimate the object area $\mu$, and later tell how we define a score; this is the same flow then as for the box center score} \dimpp{done?}

Let $\mu \left( \lVert {c_1-c_2} \rVert \right)$ be a function that estimates the logarithm of the object area (we explain how we learn this function in Sec.~\ref{sec:methodLearning}).
Based on this, for each proposal $p$ we introduce a box area score $S_{ba}$ that represents the likelihood of $p$ covering the object according to the ratio between the proposal area and the estimated object area:
\vspace{-.2cm}
\begin{equation}
\label{eq:sba}
S_{ba}(p;c_1,c_2,\sigma_{ba}) = \mathrm{e} ^ {-\frac{{ ( {a_{p}-\mu \left( \lVert {c_1-c_2} \rVert \right)}) }^2}{2\sigma_{ba}^2}}
\end{equation}
\noindent Here 
%$\mu$ returns the logarithm of the estimated object area \vitto{why $ba$?}, 
$a_p$ is the logarithm of the proposal's area, and $({a_{p}-\mu})$ indicates the log ratio between the two areas.
%\vitto{that does not work: log(a)-log(b) = log(a/b) != a/b; also: to make sense you need to symmetrize it}. 
The standard deviation $\sigma_{ba}$ controls how quickly $S_{ba}$ decreases as $a_p$ grows different from $\mu$.

Fig.~\ref{fig:methodSAexamples} shows an example of the effect of the $S_{ba}$ score on proposals of various areas. For illustration purposes, all proposals used here have a fixed aspect-ratio and are centered on the object. The score is maximal when the area of the proposal matches the estimated object area.

\mypar{Use in re-localization.}
We now use all cues in the final score function for a proposal $p$ during the re-localization step (I) of MIL step:
\vspace{-.1cm}
\begin{equation}
\label{eq:sall}
S(p) = S_{ap}(p) \cdot S_{bc}(p;c_1,c_2,\sigma_{bc}) \cdot S_{ba}(p;c_1,c_2,\sigma_{ba})
\end{equation}

\subsection{Learning score parameters}
\label{sec:methodLearning}

We exploit the clicks obtained from our qualification task on synthetic polygons to
estimate the
hyper-parameters of our model: $\sigma_{bc}$ (Eq.~\eqref{eq:sbc}), $d_{max}$
(Sec.~\ref{sec:method2click}), $\sigma_{ba}$ (Eq.~\eqref{eq:sba}) and the function $\mu$
(Eq.~\eqref{eq:sba}).
%We explain here how we learn the hyper-parameters of the model of Sec.~\ref{sec:method1click} and \ref{sec:method2click}, i.e. the $\sigma_{bc}$ of Eq.~\eqref{eq:sbc}, $d_{max}$ of Sec.~\ref{sec:method2click}, $\sigma_{ba}$ of Eq.~\eqref{eq:sba} and the function $\mu$ of Eq.~\eqref{eq:sba}.
%
%We learn these hyper-parameters without using manual bounding boxes, by exploiting the clicks on the synthetic polygons made during the qualification task (Sec.~\ref{sec:mturkQual}). 

Fig.~\ref{fig:methodPolyErrors}-left shows the distribution of the annotators' error distances during our qualification test. We estimate $\sigma_{bc}$ from this distribution.
Also, in the same figure we see that the maximum error distance is 70 pixels, hence we set $d_{max}=70$.
Fig.~\ref{fig:methodPolyErrors}-right shows the logarithm of the relative area of the synthetic polygons as a function of the distance between the two clicks. We learn the function $\mu \left( \lVert {c_1-c_2} \rVert \right)$ as a polynomial regressor fit to this data (red line in Fig.~\ref{fig:methodPolyErrors}-right). Finally, we set $\sigma_{ba}$ simply as the average error of the area estimation made by the regressor on the polygons.

\section{Experimental results}
\label{sec:results}

\subsection{Results on PASCAL VOC 2007}
\label{sec:resultsVOC}

\paragraph{Dataset.}
We perform experiments on PASCAL VOC 2007~\cite{everingham10ijcv}, which has 20 classes, 5,011 training images (trainval), and 4,952 test images. 
% We use the trainval set with accompanying image-level labels to train object detectors, and measure their performance on the test set. 
During training we only use image-level labels. 
Unlike some previous WSOL work which removes images with truncated and difficult objects~\cite{cinbis14cvpr,cinbis15pami,deselaers10eccv,russakovsky12eccv,wang15tip}, we use the complete trainval set.

%\begin{table}[t]
%\centering
%\caption{\small Details for the MS COCO training subset used for all our experiments. \dimpp{kill the table!}}
%\begin{tabular}{|c|c|c|c|} \hline
%Super-category & Category &  Train & Train(subset)\\
%\hline
%vehicle & airplane & 2243 & 552 \\
%outdoor & fire hydrant & 1205 & 328 \\
%animal & bear & 668 & 161 \\
%accesory & backpack & 3924 & 1033 \\
%sports & skateboard & 2511 & 615 \\
%kitchen & cup & 6518 & 1683 \\
%food & cake & 2080 & 557 \\
%furniture & bed & 2539 & 655 \\
%electronic & laptop & 2475 & 627 \\
%appliance & refrigerator & 1671 & 400 \\
%indoor & book & 3734 & 944 \\
%\hline	
%\hline
%\multicolumn{2}{|c|}{\textbf{Total}} & \textbf{24637} & \textbf{6303} \\
%\hline
%\end{tabular}
%\label{tab:resultCOCOsub}
%\end{table}

\mypar{Object detection model.}
As object detector we use Fast R-CNN~\cite{girshick15iccv}. Instead of Selective Search~\cite{uijlings13ijcv} we use EdgeBoxes~\cite{dollar14eccv} as proposals, as they come with an objectness
measure~\cite{alexe10cvpr} which we use inside MIL.
Unless stated otherwise, we use AlexNet~\cite{krizhevsky12nips} as the underlying CNN architecture for our method and for all compared methods.

\mypar{Evaluation.}
%Given a training set with image-level labels, our goal is to localize the object
%instances in this set and to train high quality object detectors, while minimizing human annotation
%effort. 
%We measure the trade-off between localization performance and quality of
%the object detectors against human annotation effort. % VF: unclear at this point
Given a training set with image-level labels (and possibly click annotations), our goal is to localize the object instances in this set and to train good object detectors.
We quantify localization performance on the training set with Correct Localization (CorLoc),
enabling direct comparison with WSOL methods~\cite{bilen14bmvc,bilen15cvpr,bilen16cvpr,cinbis15pami,deselaers10eccv,kantorov16eccv,russakovsky12eccv,siva11iccv,wang15tip}.
CorLoc is the percentage of images in which the bounding-box returned by the algorithm correctly localizes an object of the target class (i.e., IoU $\geq 0.5$).
We measure the performance of the trained object detector on the test set using mean average precision (mAP).
We quantify annotation effort in terms of actual human time measurements. 

%As most previous WSOL methods~\cite{bilen14bmvc,bilen15cvpr,bilen16cvpr,cinbis14cvpr,cinbis15pami,deselaers10eccv,kantorov16eccv,russakovsky12eccv,siva11iccv,song14icml,song14nips,wang15tip}, our scheme returns exactly one bounding-box per class per training image. This enables clean comparisons to previous works in terms of CorLoc on the training set.
%
%Note how at test time the object detector is capable of localizing multiple objects of the same class in the same image (and this is captured in the mAP measure).

\mypar{Compared methods.}
We compare our approach to the fully supervised alternative by training the same object detector~\cite{girshick15iccv} on the same training images, but with manually annotated bounding boxes (one per class per image, for fair comparison). 
We also compare to a modern MIL-based WSOL technique (Sec.~\ref{sec:methodMIL}) run on the same training images, but without click supervision. 

For MIL WSOL, the effort to draw bounding boxes is zero. For fully supervised learning we use the actual annotation times for ILSVRC from~\cite{su12aaai}: 35 seconds for drawing a single bounding box and verifying its quality (Sec.~\ref{sec:relwork}).
These timings are representative for PASCAL VOC, since their images are of comparable 
difficulty and quality~\cite{russakovsky15ijcv}.
%\vitto{NO: COCO is harder and we say it later}
%or higher difficulty and their annotations are of comparable quality, as discussed in~\cite{russakovsky15ijcv}. 
%The bounding-boxes in both datasets are of high quality and precisely match the object extent.

We also compare to the human verification scheme~\cite{papadopoulos16cvpr}, using their reported timings, and to various baselines.

\begin{figure}[t]
\vspace{-.4cm}
\center
\begin{tabular}{c@{}}
\includegraphics[width=1\linewidth]{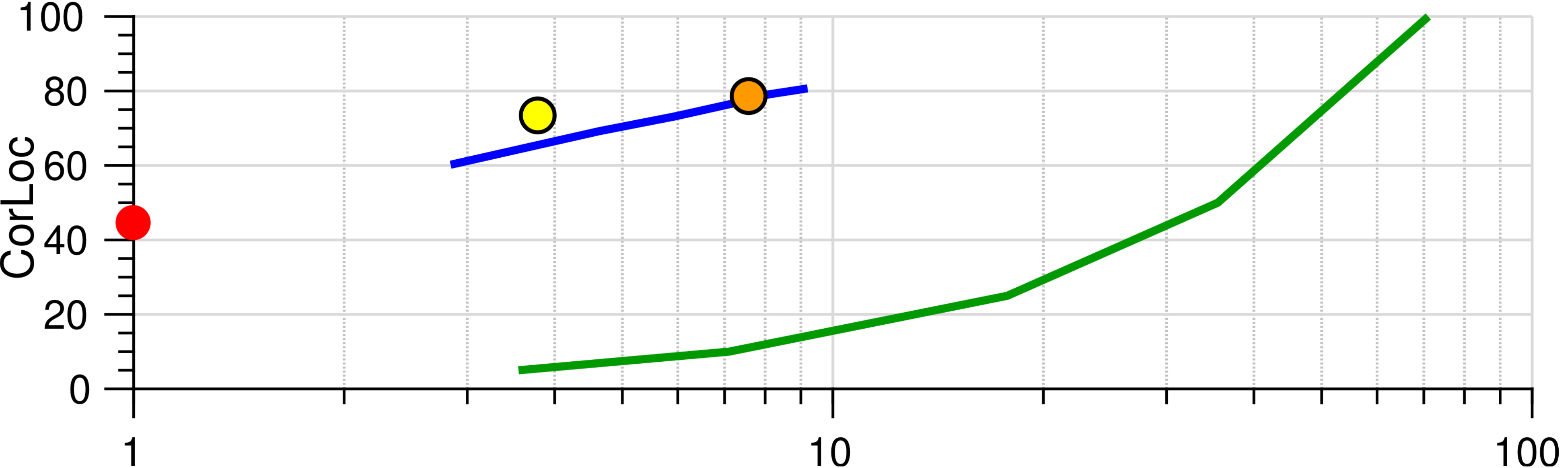} \\
\includegraphics[width=1\linewidth]{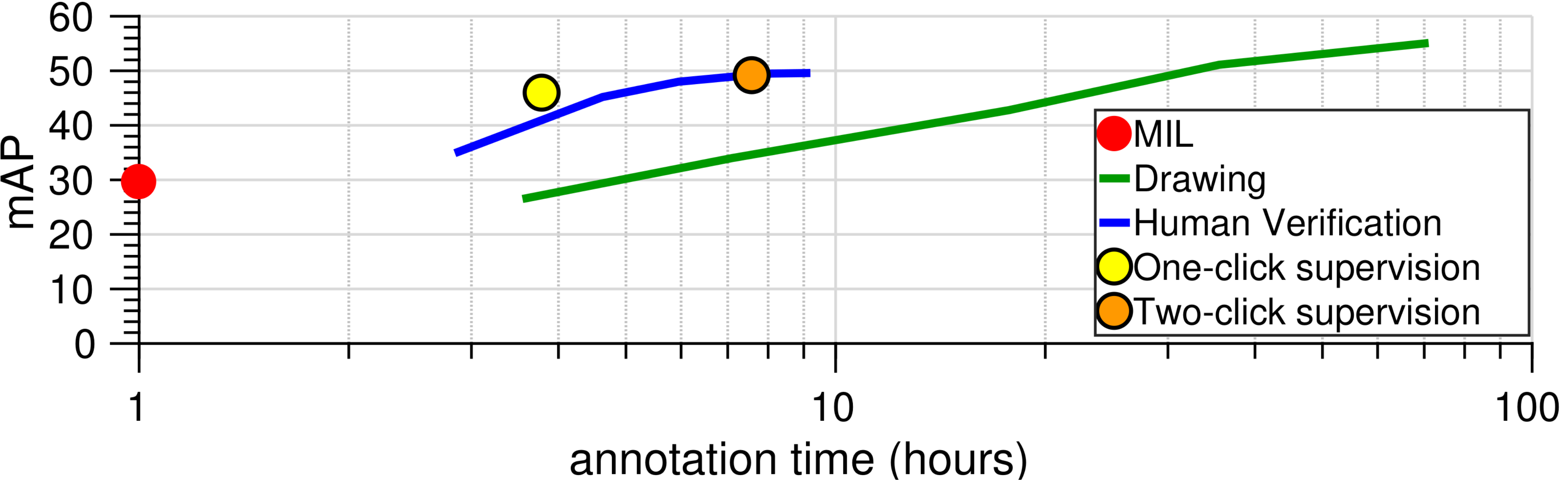} \\
\end{tabular}
\caption{\small \textbf{Evaluation on PASCAL VOC 2007.} CorLoc and mAP performance against human annotation time in hours(log-scale).}
\vspace{-.3cm}
\label{fig:resultsVOC}
\end{figure}

\mypar{Reference MIL.}
We run the reference MIL WSOL with $k=10$ folds for 10 iterations, after which it converges.
It achieves 43.4\% CorLoc on the training set. Applying two deep MIL iterations (Sec.~\ref{sec:methodMIL}) on top of this improves to 44.5\% CorLoc. The detectors produced by this approach achieve 29.6\% mAP on the test set (red dot in Fig.~\ref{fig:resultsVOC}).

%\mypar{Initializing MIL with click supervision.}
%\vitto{I think this is confusing here, especially as baselines come right next, which look similar; maybe this should go to the end? So we can start with the main result, which is about using clicks in the whole MIL}
%\vitto{we could even remove it altogther, we have too many results anyway, and this one is some weird ablation}
%The initialization of the reference MIL uses complete images as positive training examples. These initial samples have 21.6\% CorLoc.
%Instead, using one-click supervision to improve initialization as described in sec.~\ref{sec:method1click} leads to substantially better initial samples (40.4\% CorLoc).

%\mypar{One-click supervision}
% Our full model that (a) uses the $S_{bc}$ score deeply inside the MIL from the first iteration, (b) starts from a better initialization and (c) deeply retrain the network goes beyond our baselines and leads to
\mypar{One-click supervision} yields 73.3\% CorLoc. The resulting object detector yields 45.9\% mAP (yellow dot in Fig.~\ref{fig:resultsVOC}). 
Hence, at a modest extra annotation cost of only 3.8 hours we achieve an absolute improvement of +28.8\% CorLoc and +16.3\% mAP over the reference MIL.

%\mypar{Two-click supervision}
% We now evaluate our approach using two-click supervision that enables us to estimate better the object position, but also estimate the object area. While this approach
\mypar{Two-click supervision} \hspace{-2mm}  doubles the annotation time but it improves our model in two ways: (1) we can estimate the object center more accurately, and (2) we can estimate the object area based on the distance between the two clicks.  Using the two-click supervision only to improve the box center estimate $S_{bc}$ brings +0.8\% CorLoc and +0.9\% mAP over using one-click.  Including also the box area estimate $S_{ba}$ leads to a total improvement of +5.2\% CorLoc and +3.2\% mAP over one-click (78.5\% CorLoc and 49.1\% mAP, orange dot in Fig.~\ref{fig:resultsVOC}). This shows that the box area estimate contributes the most to the improvement brought by two-click over one-click supervision.

%Qualitative examples of both our one-click and two-click supervision can be seen in fig.~\ref{fig:resultsQualExamples}.

%\jasper{In general, everything has been described in the method, so you can skip telling what the method does most of the time.}

%Comparing again to the reference MIL WSOL approach, we achieve a massive jump in performance at a small extra annotation cost of 7.6 hours. (yellow dot in fig.~\ref{fig:resultsVOC}) 
%\jasper{You should only compare to 1-click and full supervision. MIL WSOL was already made obsolete by 1-click ;-)}

% The state-of-the-art weakly supervised object localization
\mypar{State-of-the-art WSOL} approaches based on AlexNet architecture~\cite{krizhevsky12nips} perform as follows.
%Bilen et al.~\cite{bilen15cvpr} achieve 43.7\% CorLoc and 27.7\% mAP,
Wang et al.~\cite{wang15tip}: 48.5\% CorLoc and 31.6\% mAP.
Cinbis et al.~\cite{cinbis15pami}: 52.0\% CorLoc and 30.2\% mAP.
Bilen et al.~\cite{bilen16cvpr}: 54.2\% CorLoc and 34.5\% mAP.
%
%\dimpp{WSOL with VGG16: ~\cite{kantorov16eccv}: 55.1\% CorLoc and 36.3\% mAP
%~\cite{bilen16cvpr}: 58.0\% CorLoc and 39.3\% mAP,}
%
Our two-click supervision outperforms all these methods with 78.5\% CorLoc and 49.1\% mAP, at a modest extra annotation cost.

%\mypar{Comparison to full supervision.}
%We now compare our approach to the standard fully supervised learning with manual bounding-boxes (green line in fig.~\ref{fig:resultsVOC}). The object detectors learned using two-click supervision achieve 49.1\% mAP, almost as good as the fully supervised ones (55.5\%). Importantly, fully supervised training needs 85 hours of annotation time when assuming 42 seconds per image. Our two-click approach instead requires only 7.6 hours reducing the human effort by a factor of 11$\times$, while our one-click approach requires only 3.8 hours, a reduction in human effort of a factor 22$\times$.

\mypar{Full supervision} \hspace{-2mm} achieves 55.5\% mAP. Our two-click supervision comes remarkably close (49.1\% mAP). Importantly, full supervision requires 71 hours of annotation time. Instead, our two-click approach requires only 7.6 hours, a reduction of~$9\times$ (or~$18\times$ for our one-click approach).
%%% it's "by a factor of n" or "by n times", not both!

\mypar{Human verification~\cite{papadopoulos16cvpr}} is shown as the blue line in Fig.~\ref{fig:resultsVOC}. 
%~\footnote{The authors of~\cite{papadopoulos16cvpr} have provided us with the results for the blue lines in fig.~\ref{fig:resultsVOC} \vitto{tricky, probably better to just blanko skip it}}.
Given the same total annotation time, our one-click method delivers higher CorLoc and mAP.
When we use two-click annotations, given the same annotation effort we match their mAP and get slightly higher CorLoc.

\mypar{Deeper CNN.}
When using VGG16~\cite{simonyan15iclr} instead of AlexNet, the fully supervised training leads to 65.9\% mAP. Our two-click model achieves 57.5\% mAP, while the reference MIL WSOL delivers 32.4\% mAP.

%\mypar{Baseline use of one-click supervision.}
%\vitto{we are short of space, I vote to kill this}
%A simple baseline using one-click supervision is to select the object proposal whose center is closest to the click position. This achieves only 26.6\% CorLoc, showing that it is not trivial to harness the click annotation information.

%\mypar{Extra ablation study.}
%\dimpp{ In order to better understand how each component of our method ($S_{ap}$, $S_{bc}$, $S_{ba}$) affects the performance, we conduct two further experiments:
%(1)Using the two-click supervision only to improve the box center estimate $S_{bc}$, discarding box area information, brings only +0.8\% CorLoc (i.e., 74.1\%) over using one-click. Instead, including the box area estimate $S_{ba}$ leads to +5.2\% CorLoc. This shows that the box area estimate contributes the most to the overall improvement brought by two-click over one-click supervision. 
%(2) Using only the box area estimate, disregarding box center information, provides a much weaker cue. It improves over the reference MIL by only +2.8\% (compared to +34.0\% brought by our full two-click method). This confirms the intuition that the position of the object yields more information than its size.}

\mypar{Effect of click accuracy.}
%We analyze here how the the accuracy of human clicks affects the localization performance of our one-click supervision model.
We compare the center-click annotations we collected (Sec.~\ref{sec:mturk}) to three alternatives:
{\it (oracle clicks)}: use the centers of the ground-truth boxes as clicks;
{\it (random clicks)}: uniformly sample a pixel inside a ground-truth box;
{\it (click-anywhere)}: we simulate a scenario where humans are instructed to click anywhere on the object, by mimicking the distribution of the publicly available click annotations of~\cite{bearman16eccv} on PASCAL VOC 2012.
We measure the distances from the centers of the ground-truth boxes to their clicks. Then we build a regressor to predict this distance based on the area of the object. Finally, we apply this regressor on VOC 2007 and displace the ground-truth object centers by the predicted distance.

For simplicity we use the alternative clicks into our one-click supervision model (Sec.~\ref{sec:method1click}) in one additional re-localization iteration at the end of the reference MIL (as opposed to using it in every iteration). For each of the three alternatives, we use the oracle best value of the parameter $\sigma_{bc}$, while for our center-click annotations we use the one learned on the synthetic polygons (sec.~\ref{sec:methodLearning}).
As a reference, when used on top of MIL this way, our center-clicks lead to 67.2\% CorLoc. Oracle clicks give an upper bound of 73.7\% CorLoc, while random clicks on the object do not improve over MIL (43.4\% CorLoc). Finally, the click-anywhere scenario achieves 55.5\% CorLoc. Interestingly, using our center-clicks leads to +11.7\% CorLoc, which shows that they convey more information.

% For reference, when used on top of MIL this way, our own crowd-sourced center-clicks lead to 67.2\% CorLoc. Oracle clicks give an upper bound of 73.7\% CorLoc, while random clicks on the object lead to a poor 24.4\% CorLoc.
% Finally, the click-anywhere scenario achieves 51.5\% CorLoc. Interestingly, this is -16.3\% CorLoc below using our center-clicks, which shows that they convey more information.

\subsection{Results on MS COCO}

\paragraph{Dataset.}
The MS COCO dataset~\cite{lin14eccv} is more difficult than PASCAL VOC, as demonstrated in~\cite{lin14eccv}, featuring smaller objects on average, and also more object classes (80).
We use exactly the same evaluation setup as for PASCAL VOC 2007 and evaluate CorLoc on the training set (82,783 images) and mAP on the val set (40,504 images).

%We report here results on a subset by picking one class per super-category: airplane, fire hydrant, bear, backpack, skateboard, cup, cake, bed, laptop, refrigerator, and book. We randomly sample 25\% of the images containing these classes, resulting in 6,303 training images,
%and we use the whole validation set as our test set (40,137 images). We use exactly the same evaluation setup as for PASCAL VOC 2007, as described in sec.~\ref{sec:resultsVOC}, and evaluate CorLoc on the training set and mAP on the test set.

%\jasper{Check numbers. You had 20k validation images!}
%Details about our training subset can be found in tab.~\ref{tab:resultCOCOsub}.

%As it is stated in~\cite{lin14eccv}, MS COCO is significantly more difficult than PASCAL VOC. To demonstrate that, in fig.~\ref{fig:resultsVOCCOCOdifficulty} we show the distribution of the square root of the relative object area for the two datasets. MS COCO contains substantially more very small objects. For example, 32\% of MS COCO objects occupy less than 1\% of the image area, while in PASCAL VOC 2007 this is true for only 5\% of the objects.

\begin{figure}[t]
\vspace{-.4cm}
\center
\begin{tabular}{c@{}}
\includegraphics[width=0.98\linewidth]{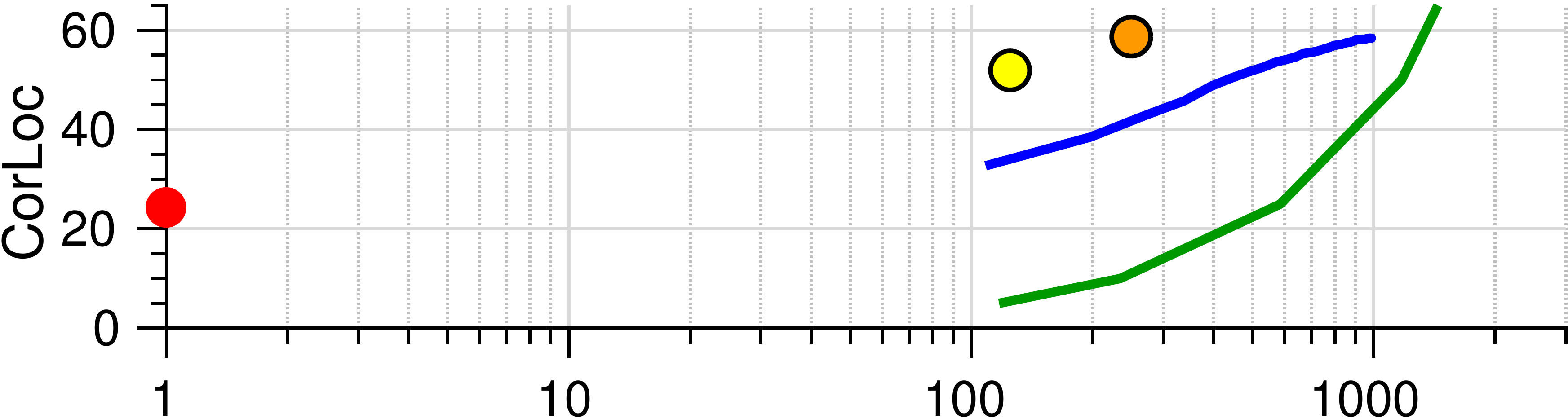} \\
\includegraphics[width=0.98\linewidth]{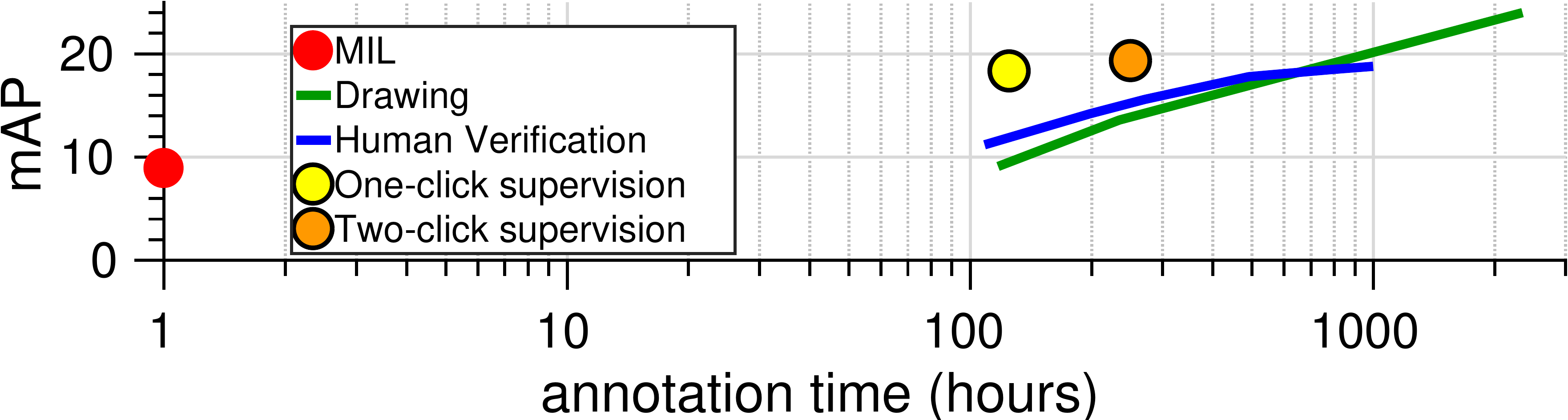} \\
\end{tabular}
\caption{\small \textbf{Evaluation on MSCOCO.} CorLoc and mAP performance against human annotation time in hours(log-scale).}
\vspace{-.3cm}
\label{fig:resultsCOCOresluts}
\end{figure}

\mypar{Reference MIL.}
The reference MIL WSOL achieves 24.2\% CorLoc and 8.9\% mAP (red dot in Fig.~\ref{fig:resultsCOCOresluts}). This is considerably lower than its performance on PASCAL VOC 2007.
%\vitto{cut if needed}
%Deselaers et al.~\cite{deselaers10eccv} also observed that the WSOL performance consistently decreases from easy datasets with mostly big objects to hard datasets with cluttered images with small objects.

%\dimpp{I also have per class analysis that it's quite interesting. Some classes completely fail (about 1\% CorLoc). I have figures that show the correlation between the CorLoc and the mean sqrtRelObjArea per class..... but there is no space!!!}

\mypar{Click supervision.}
We did not collect real click annotations for COCO, but instead simulated them. As we want to create a realistic scenario close to real annotators clicks, we did not use the centers of the available ground-truth boxes as simulated clicks. Assuming the annotator's error distance only depends on the object area, we use the findings of our error analysis on PASCAL VOC 2007 (Fig.~\ref{fig:mturkErrorArea}) to generate realistic noisy simulated clicks for COCO.

Our simulated one-click supervision 
approach 
achieves double the performance of reference MIL, reaching 51.8\% CorLoc and 18.3\% mAP (yellow dot in Fig.~\ref{fig:resultsCOCOresluts}).
Our simulated two-click supervision
approach 
goes even beyond that, with 58.6\% CorLoc and 19.3\% mAP (orange dot in Fig.~\ref{fig:resultsCOCOresluts}).
Assuming the same annotation time per click as in PASCAL VOC 2007, the total annotation time for one-click is 125 hours.
%\dimpp{cut the word approach to save a line}

\mypar{Full supervision.}
Training with full supervision requires 2,343 hours of annotation time and leads to 24.0\% mAP.

\mypar{Human verification~\cite{papadopoulos16cvpr}.}
As~\cite{papadopoulos16cvpr} do not perform experiments on COCO, we simulate their verification responses by sampling them according to the error distribution of actual humans they report on VOC. This creates a realistic simulation.
The CorLoc and mAP of this scheme can be seen in Fig.~\ref{fig:resultsCOCOresluts} (blue lines). 
%We clearly outperform this human verification scheme either by one-click or two click supervision. 
Our two-click supervision approach reaches about the same CorLoc as the simulated~\cite{papadopoulos16cvpr} (58.3\%) and it performs a bit better in terms of mAP (19.3\% vs 18.8\%). Importantly, it takes about $3.5\times$ less total annotation time.
From another perspective, given the same annotation time (250 hours), our two-click supervision approach outperforms the human verification one by +16\% CorLoc and +4\% mAP.
%\dimpp{15.7 exactly: 58.6 vs. 42.9}.
Hence, on difficult datasets with small objects our method has an edge, as the efficiency of~\cite{papadopoulos16cvpr} degrades, while the benefits of click supervision remain.

\section{Conclusions}

We proposed center-click annotation as a way of training object class detectors
and showed that crowd-sourced annotators can perform this task accurately and fast (1.9s
per object).
In extensive experiments on PASCAL VOC and MS COCO we have shown
that our center-click scheme dramatically improves over weakly
supervised learning of object detectors, at a modest additional annotation cost.
Moreover, we have shown that it reduces total annotation time by $9\times$-$18\times$ compared to manually drawing bounding boxes, while still delivering high-quality detectors. Finally, we have shown that our scheme compares favorably against a  recent method where annotators verify automatically proposed bounding boxes~\cite{papadopoulos16cvpr}.

\mypar{Acknowledgement.}
 This work was supported by the ERC Starting Grant ``VisCul''.

%\vitto{cut below?}
%We hypothesize that our framework is particularly well-suited for datasets that are harder than PASCAL, as the performance of baseline detectors will degrade, but the benefit of one-click supervision will remain. This paper presented initial evidence for this claim using simulations on MS~COCO. Studies with real annotators on a difficult dataset are the subject of future work.

{\small
\bibliographystyle{ieee}
\bibliography{/home/dim/calvin/bibtex/shortstrings,/home/dim/calvin/bibtex/vggroup,/home/dim/calvin/bibtex/calvin,/home/dim/calvin/bibtex/viscog}
}

\end{document}